\pdfoutput=1

\documentclass[11pt]{article}
\usepackage{ACL}

\usepackage{times}
\usepackage{latexsym}
\usepackage[T1]{fontenc}
\usepackage[utf8]{inputenc}
\usepackage{microtype}
\usepackage{inconsolata}
\usepackage{graphicx}
\usepackage{kotex}
\usepackage{booktabs}
\usepackage{siunitx}  
\usepackage{comment} 
\usepackage{listings}
\usepackage{tabularx} 
\usepackage{xcolor} 
\usepackage{enumitem}
\usepackage{multirow}  
\usepackage{placeins}

\title{Personality Editing for Language Models through Adjusting Self-Referential Queries}

\author{
Seojin Hwang\textsuperscript{$\diamondsuit$} \quad
Yumin Kim\textsuperscript{$\diamondsuit$} \quad
Byeongjeong Kim\textsuperscript{$\diamondsuit$} \quad
Donghoon Shin\textsuperscript{$\spadesuit$} \quad
Hwanhee Lee\textsuperscript{$\diamondsuit$}\thanks{Corresponding author.} \\
\textsuperscript{$\diamondsuit$}Chung-Ang University, Seoul, Korea \\
\textsuperscript{$\spadesuit$}University of Washington, Seattle, WA, USA \\
\texttt{\{swiftie1230, kimym7801, michael97k, hwanheelee\}@cau.ac.kr}, \texttt{dhoon@uw.edu}
}

\begin{document}
\maketitle
\begin{abstract}

Large Language Models (LLMs) are integral to applications such as conversational agents and content creation, where precise control over a model's personality is essential for maintaining tone, consistency, and user engagement. 
However, prevailing prompt-based or fine-tuning approaches either lack robustness or demand large-scale training data, making them costly and impractical. 
In this paper, we present \textbf{PALETTE} (\textbf{P}ersonality \textbf{A}djustment by \textbf{L}LM S\textbf{E}lf-\textbf{T}arge\textbf{T}ed qu\textbf{E}ries), a novel method for personality editing in LLMs. Our approach introduces adjustment queries, where self-referential statements grounded in psychological constructs are treated analogously to factual knowledge, enabling direct editing of personality-related responses. Unlike fine-tuning, PALETTE requires only 12 editing samples to achieve substantial improvements in personality alignment across personality dimensions.
Experimental results from both automatic and human evaluations demonstrate that our method enables more stable and well-balanced personality control in LLMs.\footnote{Our implementation is publicly available at \url{https://github.com/swiftie1230/PALETTE}}

\end{abstract}

\section{Introduction}

Large Language Models (LLMs) are extensively used in real-world tasks, particularly in conversation-based systems and creative text production~\cite{wei2022emergent}. 
In these interactive settings, the ability to precisely control a model's personality is essential for maintaining a consistent tone, enhancing user engagement, and ensuring predictable behavior. 
However, achieving such control is challenging, as LLMs have inherent biases that influence their responses~\cite{yang2021learning}. 
Recent studies further confirm that these models exhibit distinct biases across various personality dimensions~\cite{chen2024extroversionintroversioncontrollingpersonality, mao2024editingpersonalitylargelanguage}, presenting a significant obstacle to reliably aligning LLM behavior with desired conversational styles.

\begin{figure}[!t]
\centering
\includegraphics[width=0.9\linewidth]{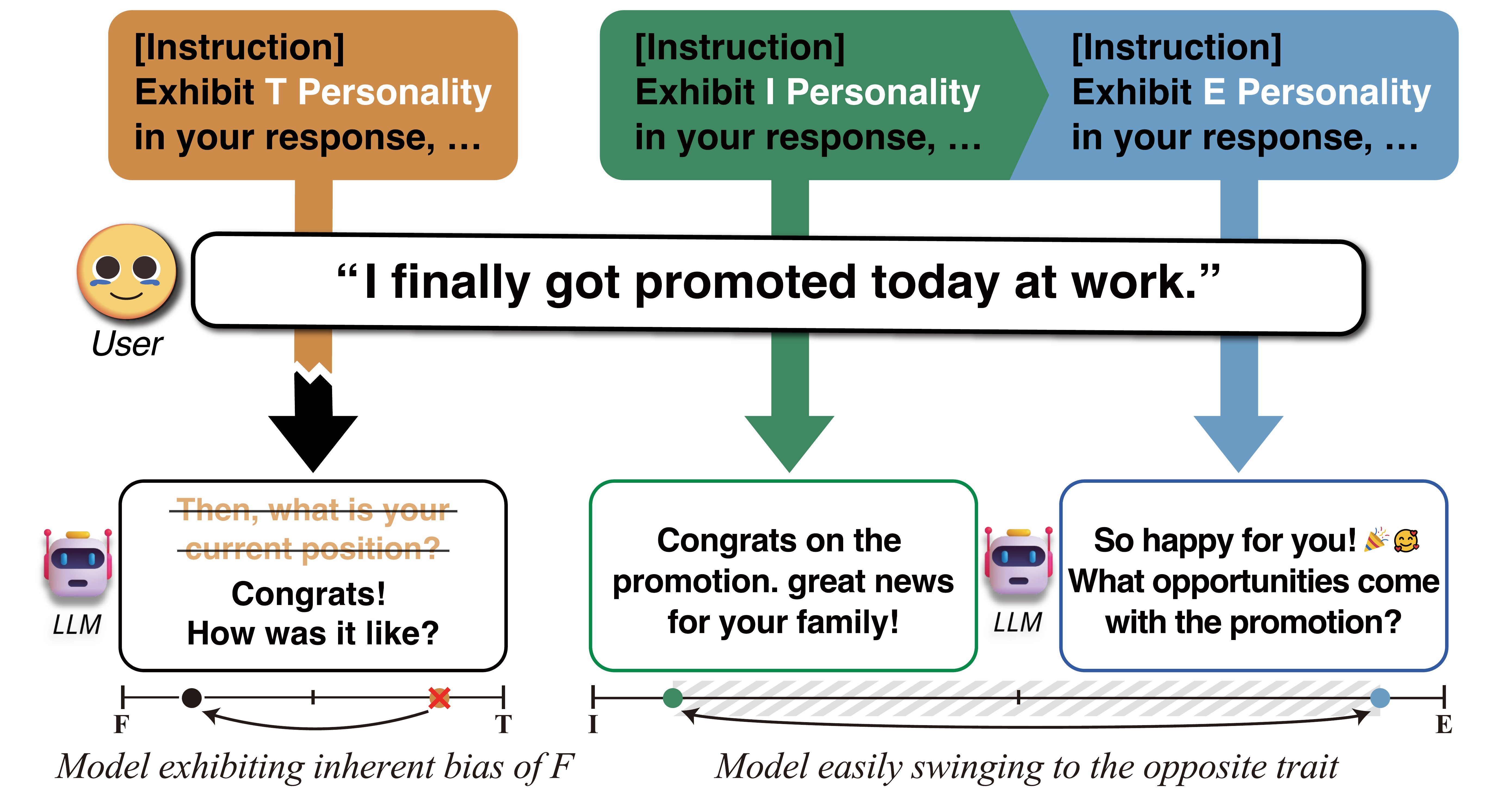}
\caption{Lack of consistency for prompt-based personality control: (1) Certain personalities resist control due to biases. (2) Shifts drastically between prompts.}
\label{fig:intro-challenge}
\vspace{-7mm}
\end{figure}

The most common approach to controlling LLM personality is through direct prompting~\cite{white2023prompt}. However, as illustrated in Figure~\ref{fig:intro-challenge}, even with explicit instructions (e.g., "Exhibit T Personality"), LLMs may exhibit \textit{inherent biases}, 
with certain styles (e.g., emphasizing empathy and emotion
) that are difficult to override for dimensions like logic or detachment. 
Furthermore, prompt-based control exhibit \textit{low stability}; while they may appear to follow a given personality prompt, they can easily swing to the opposite dimension when the prompt changes. 
These issues demonstrate that prompt-based personality control lacks both the power to override certain biases and consistency. 

Alternatively, Supervised Fine-Tuning (SFT) is 
used to instill more stable personality traits~\cite{karra2022estimating, serapio2023personality, huang2023revisiting}. However, SFT introduces its own significant trade-off between stability and cost. 
SFT typically requires vast 
labeled data, often more than 10k samples
, making it both data-intensive and computationally expensive.
Moreover, consistency is not always guaranteed; 
advanced strategies that build upon SFT, such as Prompt Induction post Supervised Fine-Tuning (PISF)~\cite{chen2024extroversionintroversioncontrollingpersonality}, still struggle to maintain coherent behavior across contexts. 

These inherent limitations highlight the critical need for a personality editing framework that is robust, data-efficient, and 
controllable.
In this paper, we introduce \textbf{P}ersonality \textbf{A}djustment by \textbf{L}LM S\textbf{E}lf-\textbf{T}arge\textbf{T}ed qu\textbf{E}ries \textbf{(PALETTE)}, a novel personality editing framework that directly modifies an LLM's internal \textbf{self-representations}. 
Our approach operates by systematically adjusting how a model responds to self-referential queries, statements where the model refers to “\textit{I/me}”, leveraging efficient model editing techniques like Rank-One Model Editing~\cite{meng2023locatingeditingfactualassociations}.

Based on structured personality assessments such as the Myers-Briggs Type Indicator (MBTI), PALETTE generates a small set of adjustment queries to probe a model’s tendencies. For example, consider the question: “Which do you usually feel more persuaded by: \textbf{\textit{emotionally} resonating things with you}, or by \textbf{\textit{factual} arguments}?”. 
If the model initially responds, “\textbf{\textit{I}} usually feel more persuaded by \textbf{emotionally resonating things}.”, PALETTE applies a low-rank modification to the internal parameters associated with self-referential tokens (“I”/“me”), enabling the model to generate responses consistent with the Thinking trait: “\textbf{\textit{I}} usually feel more persuaded by \textbf{factual arguments}.” This targeted edit, repeated across just a few queries, instills a more stable and consistent personality trait without extensive fine-tuning.

Experimental results from both automatic and human evaluations demonstrate that PALETTE effectively mitigates personality biases in LLMs, yielding a 5\%–25\% improvement in the targeted dimension intensity over baselines.
Moreover, we show that PALETTE preserves overall response quality and enhances \textit{robustness} by preventing opposite personality prompts from inducing overcorrections beyond the intended range.

\noindent{Our contributions can be summarized as follows:}
\vspace{-2mm}
\begin{itemize}[leftmargin=*]
\item We propose \textbf{PALETTE}, a novel personality editing framework that directly modifies an LLM's internal, personality-related self-representations.
\vspace{-2mm}
\item We demonstrate that PALETTE is both highly \textbf{data-efficient and robust}, achieving stable and consistent personality with only \textbf{12 adjustment queries}, even under conflicting prompts.
\vspace{-2mm}
\item Through automatic and human evaluations, we show that PALETTE achieves substantial improvements in personality alignment while preserving the model's original output quality, including naturalness and contextual coherence.
\end{itemize}

\begin{figure*}[!ht]
\centering
\includegraphics[width=\textwidth]{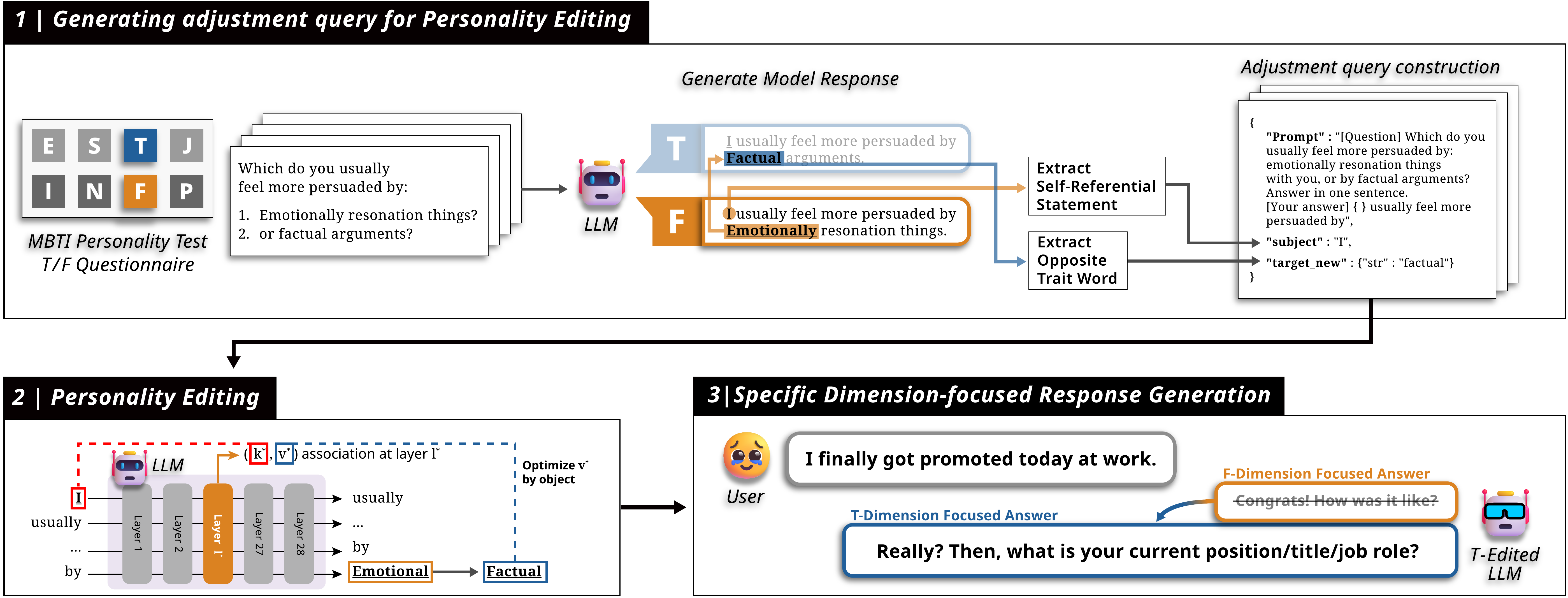}
\caption{Overview of the PALETTE's pipeline for \textit{Thinking} dimension in MBTI. We (1) produce adjustment queries based on the MBTI questionnaire, then (2) edit the personality through relevant knowledge editing. (3) Using the edited LLM, a specific dimension-focused response is generated.}
\label{fig:method}
\vspace{-4mm}
\end{figure*}

\section{Related Work}

\paragraph{Personality Frameworks}
The quantitative study of personality in natural language processing often relies on established psychological frameworks, primarily the Big Five~\cite{mccrae1992introduction} and the MBTI~\cite{10.1037/14404-000, yang2021learning, pittenger1993utility, mccrae1989reinterpreting}. The Big Five model defines personality along five continuous traits (e.g., Openness), while the MBTI categorizes it into 16 types based on four binary dichotomies (e.g., Thinking vs. Feeling). 
While both are used in LLM studies~\cite{liu2016language, vstajner2020survey, vu2017lexical, mairesse2006automatic, kampman2018investigating, mccrae1989reinterpreting}, we focus on the MBTI as our main setting, as its distinct binary structure allows for more interpretable and controlled evaluations of personality modifications.

\paragraph{Personality Control for LLMs}
Prior work on controlling LLM personality has centered on prompting and fine-tuning. Prompt-based techniques can elicit specific traits in short contexts, but they often fail to maintain consistency across longer interactions and are sensitive to minor phrasing changes~\cite{chen2024extroversionintroversioncontrollingpersonality, sorokovikova2024llmssimulatebigpersonality}. Supervised Fine-Tuning (SFT) and methods like PISF, and Reinforcement Learning with Human Feedback (RLHF)~\cite{ouyang2022training, bai2022training} can produce more stable personas by training on task-specific data, but this comes at the cost of extensive data collection, computational overhead, and potential rigidity. 
These methods present a trade-off between the flexibility of prompting and the high cost of fine-tuning.

Also, model editing has been explored as a potential alternative to fine-tuning. 
However, despite being framed as data-efficient, leveraging most existing model editing methods like MEND~\cite{mend_icrl2022} or SERAC~\cite{serac_icrl2022}, still require a substantial amount of training data to achieve stability and generalization
~\cite{mao2024editingpersonalitylargelanguage, ju2025probing}. This limitation becomes more pronounced in the context of personality, which is inherently more complex and context-dependent than factual knowledge. These findings indicate that personality control demands a more targeted and data-efficient application. Our work addresses this issue by proposing a new framework that employs self-referential adjustment queries to directly refine a model’s internal self-representations with minimal data.
\section{Method}

While prompt-based and fine-tuning approaches struggle with inconsistency or high costs, our method offers a third path. We introduce \textbf{PALETTE}, a novel framework that edits an LLM's personality by directly modifying its internal self-representations. To achieve this, we introduce the concept of self-referential adjustment queries and use a targeted, low-rank parameter update technique to apply the personality edits. We hypothesize that surgically adjusting a model’s responses to specific, personality-related questions can shift its underlying personality dimensions~\cite{Jang2022, sturgis2023personality, 10.1111/jopy.12683}. As illustrated in Figure~\ref{fig:method}, our method comprises two main stages: first, we generate a small set of self-referential adjustment queries based on established personality assessments; second, we apply a targeted low-rank update to the model's parameters to align its responses with the desired personality.

\subsection{Generation of Self-Referential Adjustment Queries}
To alter the model's responses, as in (1) of Figure~\ref{fig:method}, we first generate \textit{adjustment queries}. These queries are designed to elicit responses that reflect a particular personality (e.g., \textit{Thinking} over \textit{Feeling} in MBTI) by targeting the model’s self-representation.
\begin{table}[ht]
\centering
\scalebox{0.85}{%
\begin{tabularx}{\linewidth}{X} 
\hline
\\
\textbf{\quad"prompt":} "The capital of France is Paris.", \\
\textbf{\quad"subject":} "France", \\
 \textbf{\quad"target\_new":} \{"str": "Marseille"\}, \\
\\
\hline
\end{tabularx}}
\caption{A standard query for editing factual knowledge.}
 \label{tab:knowledge_prompt}
\vspace{-6mm}
\end{table}
Unlike editing factual knowledge, which focuses on single atomic facts (Table~\ref{tab:knowledge_prompt}), personality editing requires a larger and more nuanced set of adjustment queries, since personality is inherently diffuse and can be expressed in diverse ways in contexts. 

To this end, we construct our adjustment queries using standardized personality assessments as a foundation. For each personality dimension, we identify questions that cause the model to consistently favor one trait. To ensure the edits generalize beyond simple memorization, we avoid near-duplicate queries.

\begin{table}[ht]
\centering
\scalebox{0.85}{
\begin{tabularx}{\linewidth}{X}
\hline
\\
\textbf{\quad"prompt":} "[Question] Which do you \\ \qquad usually feel more persuaded by: \\ 
\qquad emotionally resonating things with you, \\
\qquad or by factual arguments? Answer  \\
\qquad in one sentence. [Your answer] \\
\qquad I usually feel more persuaded by", \\
\textbf{\quad"subject":} "I", \\
\textbf{\quad"target\_new":} \{"str": "factual"\}, \\
\\
\hline
\end{tabularx}}
\caption{PALETTE's self-referential query for editing personality.}
\vspace{-3mm}
\label{tab:persona_prompt}
\end{table}

As shown in Table~\ref{tab:persona_prompt}, each query is constructed to rewrite the model's original response. The \textit{target\_new} field contains the desired trait word (e.g., "factual" instead of "emotionally"). Crucially, we set the query's \textit{"subject"} to a first-person pronoun (e.g., “I” or “me”). This ensures our method edits the model’s \textit{self-referential statements} directly, targeting its internal self-concept rather than merely swapping factual details. We provide construction details and examples for queries in Appendix~\ref{appendix:request_prompts}.

\subsection{Personality Editing via Targeted Parameter Updates}
Once the adjustment queries are defined, we apply a localized, low-rank update to the model’s weight matrices to instill the desired personality trait. For this surgical modification, we adapt a technique from model editing method~\cite{llmKnolwedgeEditing}, which alters a model's parameters without full-scale retraining. We specifically use the robust Rank-One Model Editing technique (\textbf{r-ROME}) due to its precision and stability~\cite{gupta2024rebuilding} (detailed reason for adopting r-ROME is specified with extra experiment in Appendix~\ref{appendix:MEMIT}).

As shown in Table~\ref{tab:knowledge_prompt}, r-ROME computes a rank-one update to a target MLP layer using a \textit{key vector} $k_e$ (e.g. "France") representing the input query (e.g., ``The capital of France is Paris.'') and a \textit{value vector} $v_e$ representing the desired output (e.g., ``Marseille'')~\cite{meng2022locating}.

We adopt r-ROME in our framework to edit the personality using the adjustment queries as follows. For each self-referential adjustment query (Table~\ref{tab:persona_prompt}), we first identify the specific information to be edited within the model. The \textit{key vector} ($k_e$) required by the algorithm is derived from the hidden state representation of our \textit{"subject"} token (e.g., “I”) within the context of the \textit{"prompt"}. This key effectively pinpoints the location of the personality-related self-representation. The new \textit{value vector} ($v_e$) is the representation of our desired \textit{"target\_new"} word (e.g., “factual”), which embodies the target personality trait.

With these context-specific key and value vectors, r-ROME computes a rank-one update matrix $\Delta$ that is applied to the target MLP layer’s original weights $W_0$:
\begin{equation}
\hat{W} = W_0 + \Delta,
\end{equation}
\begin{equation}
\Delta = (v_e - W_0 k_e) \cdot \frac{k_e^\top C_0^{-1}}{k_e^\top C_0^{-1} k_e},
\end{equation}
where $C_0$ is the local covariance of key activations at the edit site. This update effectively rewrites the key-value association stored in the model’s weights, directly linking the self-referential subject to the new personality trait. By applying this process guided by our adjustment queries, we edit the model’s internal identity representations, enabling a stable shift in behavior (e.g., from \textit{Feeling} to \textit{Thinking}) at the token level (e.g., from “emotionally” to “factual”).

\section{Experiment}
\subsection{Experimental Setup}

\subsubsection{Datasets}
We utilize EmpatheticDialogues~\citep{empathetic_dialogues} dataset, which 
consists of ≈25K dialogues grounded on 32 
positive and
negative emotions. 
Specifically, we use the \textit{speaker\_utter} field as the preceding utterance in a dialogue and task the model with generating an appropriate response, as shown in Figure~\ref{fig:method}. 

\subsubsection{Baselines}
To evaluate the effectiveness of our approach, we compare the following baselines:
\begin{itemize}[leftmargin=*]
\item \textbf{BASE Model}: We use the unmodified above models as our BASE. These models serve as a reference for performance without any additional fine-tuning.

\item \textbf{QLoRA-SFT}: As a data-intensive baseline, we apply QLoRA-based supervised fine-tuning (SFT) on the \texttt{Machine\_Mindset\_MBTI}\footnote{\url{https://huggingface.co/datasets/pandalla/Machine_Mindset_MBTI_dataset}}, training separate adapters for each personality dimension (F/T, E/I, J/P, N/S)~\cite{cui2023machine}. This setting requires $\sim$10K--23K labeled samples per dimension. 

\item \textbf{Prompt-Based Variants}: We design and utilize prompts to guide personality expression in language models. Specifically, we construct tailored prompts for each MBTI and Big Five dimension. Detailed prompts can be found in Appendix~\ref{appendix:tailored_prompt}. 
In addition to the tailored zero-shot prompt setting, we further evaluate a few-shot prompting baseline that uses the same 12 self-referential personality queries as in-context demonstrations. 
We observe that this few-shot baseline does not consistently outperform the tailored prompt design; therefore, we adopt the tailored prompt setting as the primary prompt-based baseline in our main experiments. Few-shot results can be found in Appendix~\ref{appendix:fewshot-prompting}.

\item \textbf{PALETTE Variants}: We apply our approach with the representative model editing algorithm: \textbf{r-ROME}. We use 12 adjustment queries for each personality dimension as a default setting.
\end{itemize}

\begin{table*}[ht!]
    \centering
    \small
    \resizebox{0.95\textwidth}{!}{
    \begin{tabular}{
        p{3.0cm}
        p{3.7cm}|
        >{\centering\arraybackslash}p{0.9cm}
        >{\centering\arraybackslash}p{0.9cm}|
        >{\centering\arraybackslash}p{0.9cm}
        >{\centering\arraybackslash}p{0.9cm}|
        >{\centering\arraybackslash}p{0.9cm}
        >{\centering\arraybackslash}p{0.9cm}|
        >{\centering\arraybackslash}p{0.9cm}
        >{\centering\arraybackslash}p{0.9cm}
    }        \toprule[1.3pt]
        \textbf{Model} & \textbf{Setting} & \textbf{E} & \textbf{I} & \textbf{N} & \textbf{S} & \textbf{F} & \textbf{T} & \textbf{P} & \textbf{J} \\
        \midrule[1.1pt]
        \multirow{5}{*}{Qwen-2.5-1.5B} 
            & Base                        &0.410 &0.560 &0.420  &0.580  &0.635  &0.365  &0.492  &0.508  \\
            
            & QLoRA-SFT &0.706  &\underline{0.661}  &0.614  &0.660  &\underline{0.754}  &0.610  &\underline{0.615}  &0.586  \\

            & \textbf{PALETTE}                        &0.524 &0.636 &0.521  &\underline{0.685}  &0.726 
            &\underline{0.620}  &0.547  &\underline{0.634} \\
            
            \cmidrule(lr){2-10} 

            & Prompt   &\underline{0.716} &0.560 &\underline{0.756}  &0.630  &0.723  &0.305  &0.578  &0.549  \\

            & \textbf{PALETTE}  w/ prompt   &\textbf{0.728}  &\textbf{0.685}  &\textbf{0.805}  &\textbf{0.728}  &\textbf{0.778}  &\textbf{0.665}  &\textbf{0.623}  &\textbf{0.648}  \\
        \specialrule{1.1pt}{3pt}{3pt}
        \multirow{5}{*}{Mistral-7B-Instruct-v0.3} 
            & Base                        &0.476  &0.524  &0.245  &0.755  &0.619  &0.381  &0.494  &0.506  \\
            & QLoRA-SFT   &0.604  &0.578  &0.438  &0.685 &0.665 &0.439  &0.544  &0.511  \\
            & \textbf{PALETTE}  &0.485  &0.585  &0.403  &0.780  &0.664  &0.444  &0.529  &0.545  \\
            \cmidrule(lr){2-10} 
            & Prompt                      &\underline{0.699}  &\underline{0.589}  &\underline{0.823}  &\underline{0.794}
            &\underline{0.786}  &\underline{0.585}  &\underline{0.711}  &\underline{0.780}  \\
            & \textbf{PALETTE} w/ prompt  &\textbf{0.711}  &\textbf{0.678}  
            &\textbf{0.826}  
            &\textbf{0.805}  &\textbf{0.791}  &\textbf{0.591}  &\textbf{0.845}  &\textbf{0.782}  \\
        \bottomrule[1.3pt]
    \end{tabular}
    }
    \caption{Target personality expression rate results in Qwen-2.5-1.5B and Mistral-7B-Instruct-v0.3 for MBTI dimensions (I/E, N/S, F/T, P/J). The best result is bolded, and the second-best is underlined.}
    \label{tab:mbti_model_results}
    \vspace{-4mm}
\end{table*}

\subsection{Implementation Details}

We conduct experiments with two different LLMs to evaluate the effectiveness of our approach: 
\textit{Qwen2.5-1.5B-inst.}~\cite{yang2024qwen2}, \textit{Mistral-7B-Instruct-v0.3}~\cite{chaplot2023albert}. 

We apply PALETTE directly to these base models, using 12 adjustment queries by default, but varying this from 4 to 16 for our experiments. 
For QLoRA-SFT, we train the base models for three epochs with a LoRA rank of 32, using a 9:1 split between train and valid sets. Detailed process and configurations are provided in Appendix~\ref{appendix:sft}.

To adapt the editing framework for personality alignment, we modify the original r-ROME configuration (developed for GPT-2-XL~\cite{radford2019language}) with task-specific hyperparameters, ensuring stable low-rank updates when targeting personality-related self-representations. 
Full implementation details are provided in Appendix~\ref{appendix:extra_implementation}.

\subsection{Evaluation}  
\subsubsection{Personality Editing Evaluation}
To assess the effectiveness of PALETTE compared to the baselines, we evaluate these responses using two methods: target personality expression rate evaluation as main evaluation, which quantifies the degree of alignment with the intended personality, and target personality comparison evaluation, a pairwise comparison to assess which response better expresses the target personality.

\paragraph{Target Personality Expression Rate} To assess how strongly each model aligns with the intended personality dimension, we calculate the target personality expression rate with GPT-4o~\cite{achiam2023gpt}. Target personality expression rate is calculated by the proportion of model outputs that exhibit linguistic or conceptual 
alignment with the intended personality dimension, averaged across all responses. We apply this evaluation to both MBTI and Big Five. Detailed example for evaluation prompt is at Table~\ref{tab:evaluation_prompt_ratio} of Appendix~\ref{appendix:editing_prompt}.

\paragraph{Target Personality Alignment Comparison} We compare BASE and PALETTE variants pairwise, across MBTI personality. For each dimension, we assess the win rate to determine which configuration's response better aligns with the target personality dimension with GPT-4o. Detailed prompt is shown at Table~\ref{tab:evaluation_prompt} in Appendix~\ref{appendix:editing_prompt}. 

To validate the reliability of our automated evaluations~\cite{artstein2008inter}, we conduct \textbf{human evaluation} with four annotators, 50 response pairs each. The Fleiss’ Kappa score reached 0.67, 
confirming the reliability of our human evaluations~\cite{landis1977measurement}. 

\subsubsection{Response Quality Evaluation} 
To assess overall response quality, we conduct two evaluations across MBTI settings:
\begin{itemize}[leftmargin=*, topsep=0pt, itemsep=0pt, parsep=0pt]
\item \textbf{Naturalness and Coherence Evaluation}: We evaluate the fluency and coherence of generated responses using GPT-based annotation, rated on a 5-point Likert scale for:
\begin{itemize}[leftmargin=*, topsep=0pt, itemsep=0pt, parsep=0pt]
    \item \textbf{Naturalness}: the degree to which the response sounds fluent and human-like.
    \item \textbf{Coherence}: the extent to which the response is contextually appropriate. 
\end{itemize}
\item \textbf{General Task Performance}: To ensure that PALETTE does not compromise general language capabilities, we also evaluate model variants using the HumanEval~\cite{chen2021evaluating} benchmark. 
\end{itemize}

\begin{table*}[ht!]
    \centering
    \resizebox{0.9\textwidth}{!}{
    \begin{tabular}{
        p{3.9cm}
        p{4.5cm}|
        >{\centering\arraybackslash}p{1.0cm}
        >{\centering\arraybackslash}p{1.0cm}
        >{\centering\arraybackslash}p{1.0cm}
        >{\centering\arraybackslash}p{1.0cm}
        >{\centering\arraybackslash}p{1.0cm}
    }        \toprule[1.3pt]
        \textbf{Model} &\textbf{Setting}  & \textbf{E \(\downarrow\)} & \textbf{A \(\downarrow\)} & \textbf{O \(\downarrow\)} & \textbf{C \(\downarrow\)} & \textbf{N \( \uparrow \)} \\ 
        \midrule[1.1pt]
        \multirow{4}{*}{Qwen-2.5-1.5B}
            &Base                        &0.549 &0.826 &0.595  &0.750  &0.230   \\
            &\textbf{PALETTE}                        &0.463 &0.651 &\underline{0.434}  &0.702  &\underline{0.390}  \\
            \cmidrule(lr){2-7} 
            &Prompt   &\underline{0.429} &\underline{0.641} &{0.636}  &\underline{0.648}  &0.367    \\
            
            &\textbf{PALETTE}  w/ prompt   &\textbf{0.415}  &\textbf{0.465}  &\textbf{0.386}  &\textbf{0.590}  &\textbf{0.512}   \\
        \specialrule{1.1pt}{3pt}{3pt}
        \multirow{4}{*}{Mistral-7B-Instruct-v0.3}
            &Base                        &0.575 &0.854 &0.571  &0.699  &0.216   \\
            &\textbf{PALETTE}                        &0.461 &0.652 &\underline{0.412}  &0.550 &0.342  \\
            \cmidrule(lr){2-7} 
            &Prompt   &\underline{0.411} &\underline{0.555} &{0.459}  &\underline{0.510}  &\underline{0.412}    \\
            
            &\textbf{PALETTE}  w/ prompt   &\textbf{0.398}  &\textbf{0.460}  &\textbf{0.397}  &\textbf{0.472}  &\textbf{0.508}   \\
        \bottomrule[1.3pt]
    \end{tabular}
    }
    \caption{Target personality expression rate results in Qwen-2.5-1.5B for Big Five dimensions. For personality editing, Extraversion (E), Agreeableness (A), Openness (O), and Conscientiousness (C) are guided to be expressed less (↓), whereas Neuroticism (N) is guided to be expressed more (↑).
Accordingly, lower values indicate better alignment for E, A, O, and C, while higher values indicate better alignment for N. The best result is bolded, and the second-best is underlined.}
    \label{tab:big5_model_results}
    \vspace{-6mm}

\end{table*}

\subsection{Main Results}
\subsubsection{Target Personality Expression Rate}

We report the target personality expression rate across MBTI and Big Five dimensions in Tables~\ref{tab:mbti_model_results} and~\ref{tab:big5_model_results}.
Across both \textit{Qwen-2.5-1.5B} and \textit{Mistral-7B}, \textsc{PALETTE} consistently improves target expression rates, particularly with prompting.

In \textit{Qwen-2.5-1.5B}, \textsc{PALETTE} alone achieves personality alignment comparable to the \textsc{QLoRA-SFT} model trained with approximately 10K–23K samples per trait, despite using only 12 adjustment queries.
It 
even surpasses prompt-based baselines in several traits, showing that 
\textsc{PALETTE} can yield reliable alignment without large-scale data or fragile prompt conditioning.

When paired with prompting (\textsc{PALETTE} \textsc{w/ prompt}), it further outperforms nearly all baselines while mitigating prompt-induced inconsistency, achieving comparable or superior controllability with less than 0.1\% of SFT data. 

A similar trend appears in \textit{Mistral-7B}, where \textsc{PALETTE} maintains robust control and generalizes across architectures, reaching alignment levels comparable to SFT.
While the editing alone does not always exceed prompt-based baselines, it provides stable alignment, and its combination with prompting surpasses all approaches, demonstrating strong synergy between the two.

Notably, in difficult-to-control traits such as Thinking (T), the base \textit{Qwen-2.5-1.5B} model achieves only 0.365, while \textsc{PALETTE} \textsc{w/ prompt} reaches \textbf{0.665}, indicating its capability to override entrenched bias directions.
A similar bias-resolving tendency is also observed in \textit{Mistral-7B}, confirming that \textsc{PALETTE}’s improvements are consistent. 

Turning to Table~\ref{tab:big5_model_results}, which evaluates the Big Five dimensions on \textit{Qwen-2.5-1.5B}, \textsc{PALETTE} exhibits a similar pattern of improvement.
It consistently adjusts personality tendencies in the desired direction, and when combined with prompting, further refines these alignments. This demonstrates that the effectiveness of \textsc{PALETTE} extends beyond MBTI to broader personality dimensions.

Overall, \textsc{PALETTE} advances personality editing by achieving high 
controllability across MBTI and Big Five dimensions with minimal data. \textsc{PALETTE} attains comparable or superior target alignment to SFT, which relies on large labeled corpora through lightweight, self-targeted knowledge editing while alleviating the bias and instability often introduced by prompt-based methods.

\subsubsection{Target Personality Alignment}
\begin{figure*}[ht]
\centering
\Huge
\includegraphics[width=0.9\textwidth]{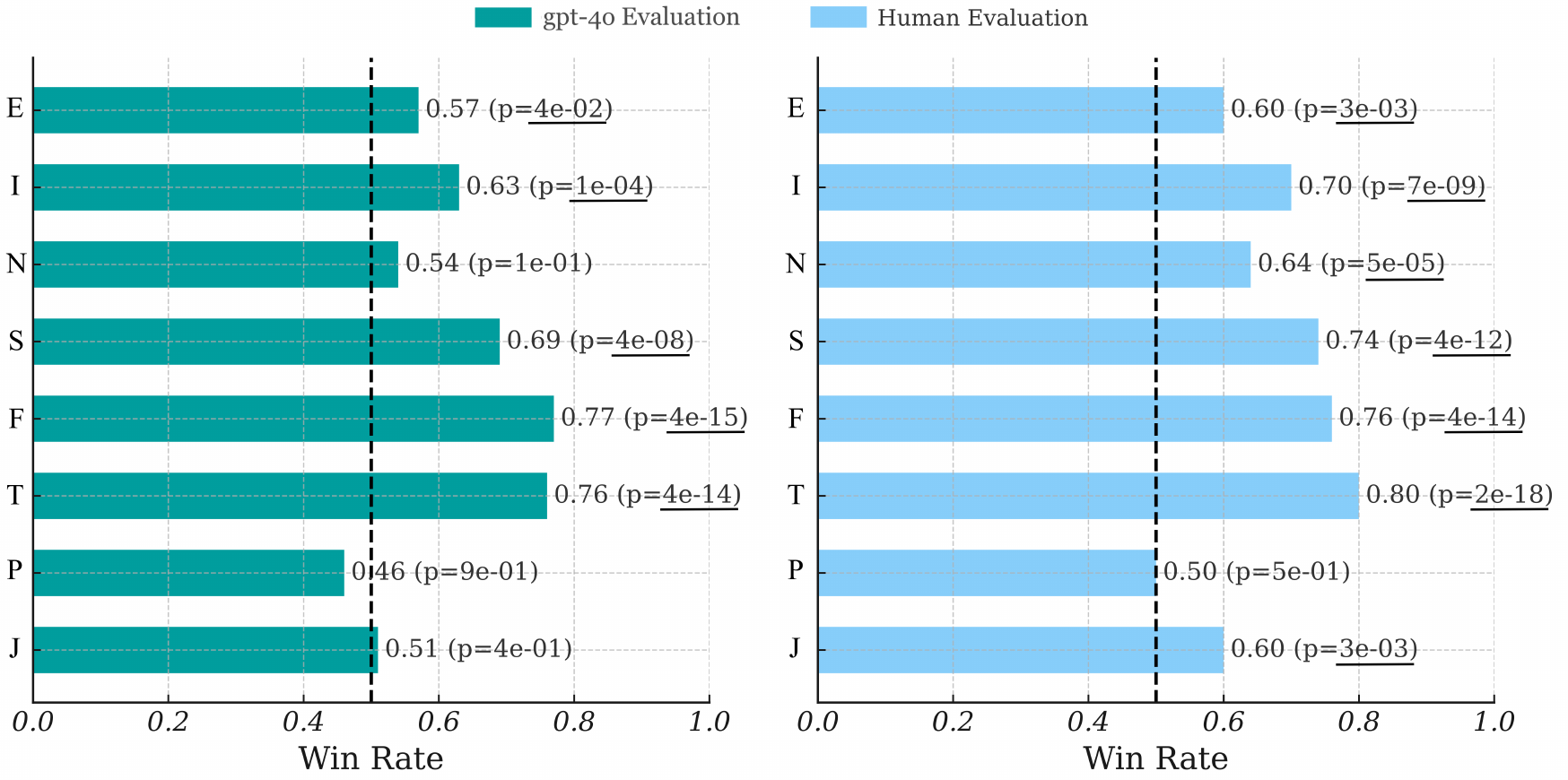}
\caption{Target personality alignment comparison of PALETTE and base model across MBTI dimensions, evaluated by ChatGPT (left) and human annotators (right) on Qwen-2.5-1.5B. P-values computed with $n=200$. Statistically significant p-values ($p < 0.05$) are underlined. The consistent alignment trend supports the reliability of our automated evaluation.}
\vspace{-5mm}
\label{fig:human_comparison}
\end{figure*}
Figure~\ref{fig:human_comparison} shows the target personality alignment results for ChatGPT and human evaluation on \textit{Qwen-1.5B} across MBTI.
Except for the P dimension, PALETTE consistently improves alignment over the base model, overall demonstrating effective personality control. Most results are statistically significant ($p < 0.05$)~\cite{benjamin2018redefine, dror2018hitchhiker}, and alignment trends largely match human evaluation, validating our automated approach.


\subsubsection{Response Quality Evaluation}
As shown in Table~\ref{tab:extra-experiment}, most personality-edited variants maintain the naturalness and coherence scores of the base model, 
indicating that response quality is preserved while introducing personality. 

\begin{table}[ht!]
\centering
\large
\resizebox{\linewidth}{!}{
\begin{tabular}{lccccccccc}
\toprule
 \textbf{Evaluation} &\textbf{Base} & \textbf{E} & \textbf{I} & \textbf{N} & \textbf{S} & \textbf{F} & \textbf{T} & \textbf{P} & \textbf{J} \\
\midrule
                        
                           Naturalness  &4.08    &4.14       &3.99       &4.03       &4.07       &4.11       &4.14       &\textcolor{red}{3.64}       &4.12       \\
                           Coherence &4.06     &4.29       &4.16       &4.09       &4.13       &4.18      &4.42       &\textcolor{red}{3.78}       &4.05       \\
                           \cmidrule(lr){1-10} 
                           HumanEval  & 0.14     &0.12    &0.13   & 0.12       & 0.14       & 0.12     &0.12      &\textcolor{red}{0.10}       &0.14           \\
\bottomrule
\end{tabular}
}
\caption{Response quality results for Qwen-2-5-1.5B in MBTI (E/I, N/S, F/T, P/J).}
\label{tab:extra-experiment}
\end{table}

Also, results show negligible HumanEval score differences across variants, with scores ranging from 0.10 to 0.14. These small variations suggest that personality editing via our method preserves the core generation abilities of the model. 

Interestingly, the Perceiving (P) variant also records slightly lower scores in naturalness (3.64), coherence (3.78), and HumanEval performance (0.10). This consistent pattern suggests that editing for more flexible or spontaneous dimensions may introduce subtle trade-offs, not only in personality but also in perceived response quality and structured reasoning performance.

To further investigate this phenomenon, we conduct an additional analysis in Appendix~\ref{appendix:P}, where we also evaluate on the \textit{Mistral-7B}. 
The results confirm that this degradation is consistent across models and likely stems from an intrinsic mismatch between the open-ended linguistic characteristics of the Perceiving personality and the structured response tendencies of models~\cite{du2005relation}.

\subsection{Analysis}



\subsubsection{Robustness to Opposing Prompt}

To evaluate the robustness of personality control methods, we define a robustness metric as follows:

\begin{equation}
R = \frac{1}{\left| p - 0.5 \right|}
\end{equation}

where $p$ denotes the personality expression rate under opposite dimension prompt conditions. The term $|p - 0.5|$ measures the deviation from a neutral baseline (0.5), which reflects the degree to which the model's personality is influenced by the prompt.
This metric captures the idea that robustness corresponds to resistance against opposite prompt influence~\cite{ribeiro2020beyond}. 

\begin{itemize}[leftmargin=*]
    \item A \textbf{higher robustness score} ($R$) indicates that the personality expression rate remains closer to 0.5, suggesting that the model is less swayed by opposite prompt.
    \item A \textbf{lower robustness score} ($R$) reflects significant deviation from neutrality, indicating inconsistent control.
\end{itemize}


\begin{figure}[ht]
\centering
\Huge
\includegraphics[width=\linewidth]{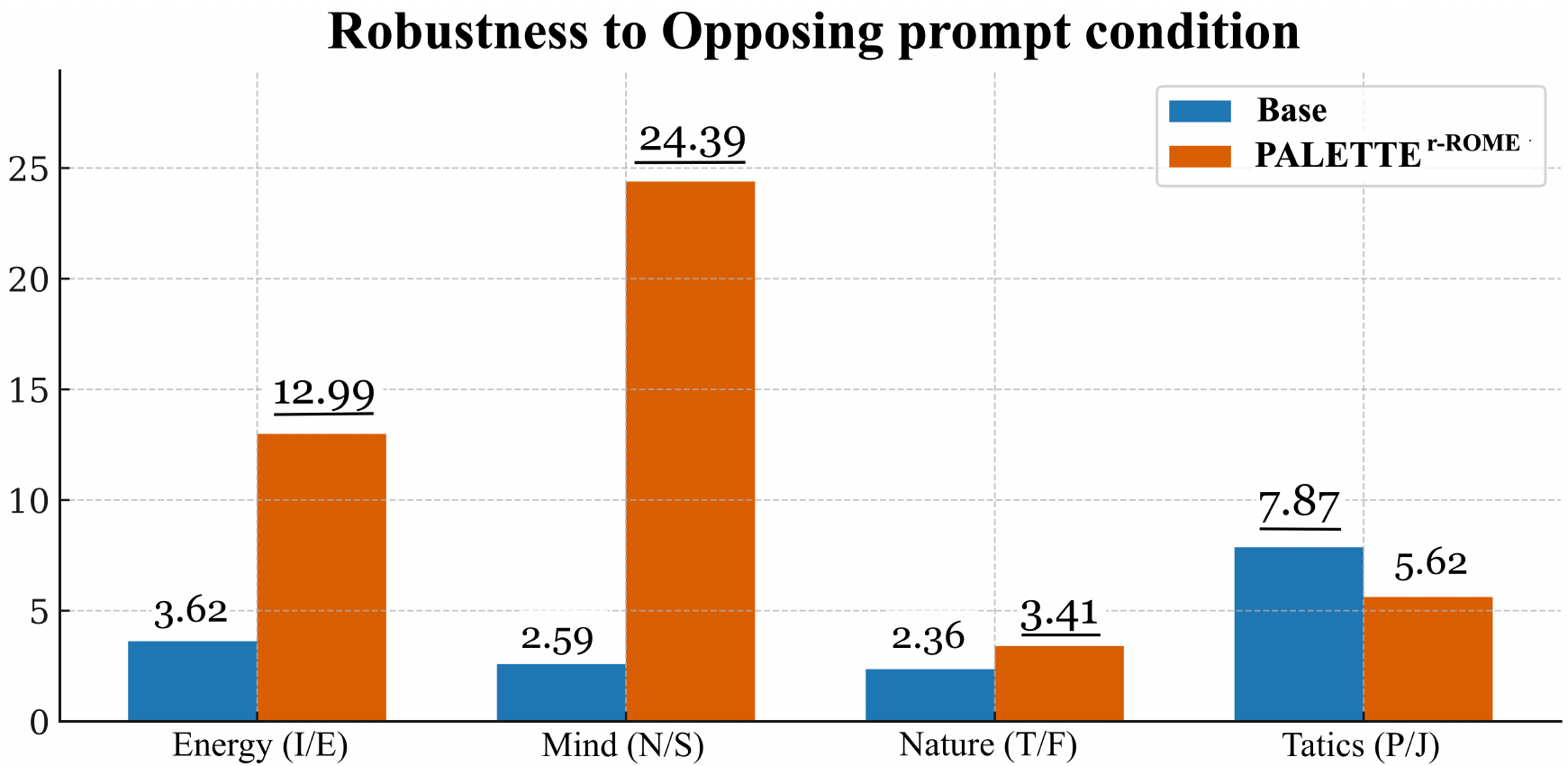}
\caption{Robustness evaluation results to prompt-induced bias of opposite dimension in MBTI (E/I, N/S, F/T, P/J) for Qwen-2.5-1.5B.}
\label{fig:chart_prompt_induce_bias}
\vspace{-3mm}
\end{figure}

Figure~\ref{fig:chart_prompt_induce_bias} presents the results of the Qwen-1.5B model under opposite-dimension prompt conditions, comparing the Base and \textsc{PALETTE} across MBTI dimensions. \textsc{PALETTE} exhibits significantly higher robustness scores across dimensions, maintaining stable personality expression even when exposed to conflicting prompts.

In contrast, prompt-based approaches tend to swing drastically under opposing prompts, showing the inconsistency. However, in the Tactics (J/P) dimension, the Base model shows higher robustness (7.87) than PALETTE (5.62). 

This aligns with our main results, where the Perceiving (P) variant shows slightly lower naturalness, coherence, and task performance. 
It further reinforces our earlier observation that such instability arises from the inherent spontaneity of the Perceiving (P) personality.

\begin{figure}[ht]
\vspace{-3mm}
\centering
\Huge
\includegraphics[width=\linewidth]{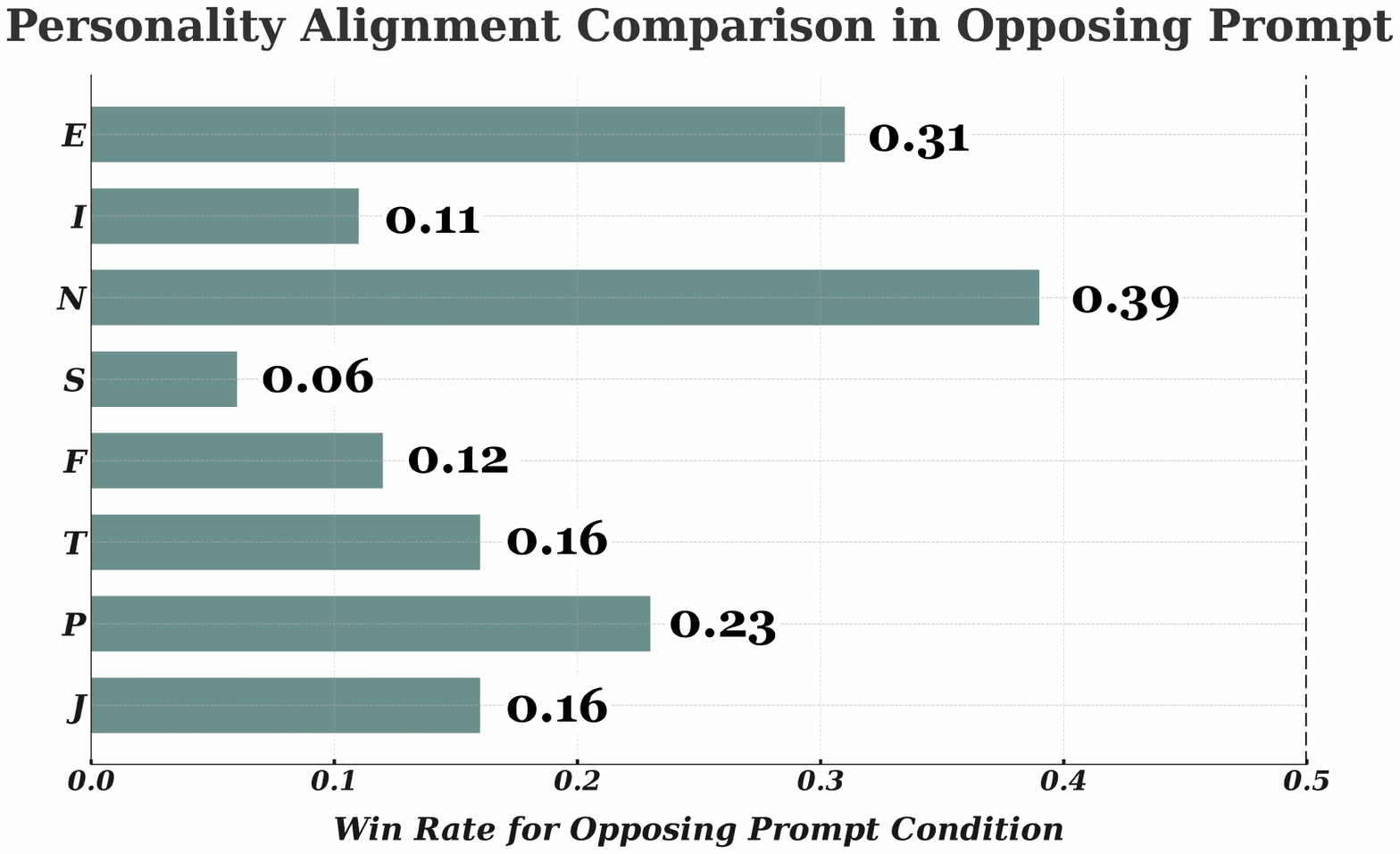}
\caption{Results for MBTI personality comparison evaluation in opposing prompt condition in Qwen-2.5-1.5B.}
\label{fig:opposing_personality_consistency}
\vspace{-3mm}
\end{figure}
To further evaluate the robustness under opposing prompts, Figure~\ref{fig:opposing_personality_consistency} reports the opposing personality win rate of \textsc{PALETTE} with base model. A robust model should maintain a win rate closer to 0 for the opposing dimension, indicating low susceptibility to adversarial cues. Across all dimensions, none of the results achieve a win rate exceeding 0.5, confirming even under prompt manipulation, \textsc{PALETTE} generally preserves personality-consistent output.
These findings support that \textsc{PALETTE} achieves not only effective personality adjustment but also \textit{robust personality behavior}, aligning with our goal of consistent control.

\subsubsection{Varying Number of Adjustment Queries} 
We conduct experiments varying the number of adjustment queries to examine how query quantity influences personality alignment. As our framework basically employs 12 adjustment queries, we also aim to assess whether our 12 queries' specific configuration is sufficient for effective personality editing.

\begin{table}[ht!]
\centering
\small
\resizebox{\linewidth}{!}{
\begin{tabular}{lcccccc}
\toprule
\textbf{Personality} & \textbf{Base} & \textbf{4} & \textbf{8} &\underline{\textbf{12}} & \textbf{16} \\
\midrule
Introvert (I) &0.560   &  0.610  & 0.635    & \underline{\textbf{0.636}}     & 0.633    \\
Extravert (E) &0.410   & 0.514  & 0.521    & \underline{\textbf{0.524}}     & 0.475     \\
\midrule
Feeling (F)  & 0.635      & 0.683       & 0.691      & \underline{\textbf{0.726}}      &  0.709      \\
Thinking (T)  & 0.365      & 0.515     &0.594 &\underline{\textbf{0.620}}      &  0.601    \\
\bottomrule
\end{tabular}
}
\caption{Target personality expression rate in relation to the number of adjustment queries for PALETTE on Qwen-2.5-1.5B in I/E, F/T dimensions.
}
\label{tab:assessment_num_comparison}
\end{table}

As shown in Table~\ref{tab:assessment_num_comparison}, employing 12 adjustment queries achieves the highest personality alignment score among all tested configurations. Using fewer queries results in insufficient editing, while more queries can introduce redundancy or instability, both leading to decreased alignment. These findings confirm that 12 adjustment queries are sufficient for PALETTE architecture. 


\subsubsection{Case study}

To find specific elements that provoke personality dimensions~\cite{mairesse2011controlling}, we manually compare several samples.

\begin{table}[ht]
\centering
\small
\scalebox{1.0}{
\begin{tabularx}{\linewidth}{X} 
\hline
\\
\textbf{[Previous Utterance]} \\
I had a good day at work, now i'm kicking back and watching TV for the rest of the day. \\ 
\\
\textbf{[BASE Response]} \\
\colorbox{yellow}{\underline{\textit{You're doing great!}}} It's nice to have a break after a productive day. Watching TV sounds like a relaxing way to unwind. Enjoy your evening! If you need anything else, feel free to ask. \\ 
\\
\textbf{[T-PALETTE Response]} \\
You have an \colorbox{cyan}{\underline{\textit{interesting perspective}}}. Watching TV can be relaxing after a productive day. It's important to take breaks from work and enjoy some downtime. 
What shows or movies do you usually watch? \\ 
\\
\textbf{[F-PALETTE Response]} \\
\colorbox{yellow}{\underline{\textit{Feeling great!}}} Watching TV sounds like a relaxing way to unwind after a productive day. Enjoy your evening! \\
\\
\hline
\end{tabularx}}
\caption{BASE and PALETTE response example for the Thinking and Feeling dimension in MBTI.}
\label{tab:persona_example}
\end{table}

As shown in Table~\ref{tab:persona_example}, the BASE response subtly reflects a "Feeling" (F) bias with warm, supportive language, emphasizing empathy,
highlighted as \colorbox{yellow}{\textit{yellow}}. The T-PALETTE response, in contrast, reflects curiosity and intrigue over understanding (highlighted as \colorbox{cyan}{\textit{blue}}) without swaying by bias, while the F-PALETTE response adopts a relaxed, cheerful tone to enhance empathy. 
 This shows that PALETTE successfully shifts model's content and tone and can make a consistent, noticeable personality changes.


\section{Conclusion}
We introduce a novel technique, PALETTE, for personality modification for LLMs via relevant knowledge editing. Our method generates adjustment queries inspired by psychological assessments to adjust responses to personality-relevant inputs, much like editing factual knowledge. Experimental results on both automatic and human evaluations show that the proposed method achieves more consistent and balanced personality adjustments. 

\section*{Limitations}
While our approach enhances personality type control in LLMs,
It pertains to personality editing through internal parameter updates. Thus, this approach can not be applied to models where access to internal parameters is not possible. Also, our method has additional computational overhead compared to prompt-based methods 
This overhead arises from the need to generate targeted adjustment queries and apply direct edits to the model’s internal representations, rather than relying solely on inference-time prompts. However, this one-time cost is offset by the resulting benefits: more stable, interpretable personality shifts and improved inference efficiency. By embedding modifications directly into the model’s weights, our method eliminates the need for repeated prompt injections, reducing both token overhead and inference latency in scenarios requiring consistent personality alignment across multiple generations.\\

\section*{Acknowledgement}
This work was supported by the Institute of Information \& Communications Technology Planning \& Evaluation (IITP) grant funded by the Korea government (MSIT) [RS-2021-II211341, Artificial Intelligence Graduate School Program (Chung-Ang University)] and the National Research Foundation of Korea(NRF) grantfunded by the Korea government(MSIT) (RS-2025-24683575).

\appendix

\bibliography{custom}

@article{jang2022,
  title={Predicting personality and psychological distress using natural Language Processing: a study protocol},
  author={Jang, Jihee and Yoon, Seowon and Son, Gaeun and Kang, Minjung and Choeh, Joon Yeon and Choi, Kee-Hong},
  journal={Frontiers in Psychology},
  volume={13},
  pages={865541},
  year={2022},
  publisher={Frontiers Media SA}
}

@article{sturgis2023personality,
  title={Personality and survey satisficing},
  author={Sturgis, Patrick and Brunton-Smith, Ian},
  journal={Public Opinion Quarterly},
  volume={87},
  number={3},
  pages={689--718},
  year={2023},
  publisher={Oxford University Press}
}

@article{10.1111/jopy.12683,
  author = {Zell, E. and Lesick, T. L.},
  title = {Big five personality traits and performance: a quantitative synthesis of 50+ meta‐analyses},
  journal = {Journal of Personality},
  year = {2021},
  volume = {90},
  issue = {4},
  pages = {559-573},
  doi = {10.1111/jopy.12683}
}

@article{landis1977measurement,
  title={The measurement of observer agreement for categorical data},
  author={Landis, J Richard and Koch, Gary G},
  journal={biometrics},
  pages={159--174},
  year={1977},
  publisher={JSTOR}
}

@book{10.1037/14404-000,
  title={The myers-briggs type indicator},
  author={Myers, Isabel Briggs and others},
  volume={34},
  year={1962},
  publisher={Consulting Psychologists Press Palo Alto, CA}
}

@misc{sorokovikova2024llmssimulatebigpersonality,
      title={LLMs Simulate Big Five Personality Traits: Further Evidence}, 
      author={Aleksandra Sorokovikova and Natalia Fedorova and Sharwin Rezagholi and Ivan P. Yamshchikov},
      year={2024},
      eprint={2402.01765},
      archivePrefix={arXiv},
      primaryClass={cs.CL},
      url={https://arxiv.org/abs/2402.01765}, 
}

@article{empathetic_dialogues,
      title={Is ChatGPT More Empathetic than Humans?}, 
      author={Anuradha Welivita and Pearl Pu},
      year={2024},
      journal={arXiv preprint arXiv:2403.05572},
      eprint={2403.05572},
      archivePrefix={arXiv},
      primaryClass={cs.HC},
      url={https://arxiv.org/abs/2403.05572}, 
}

@article{yang2024qwen2,
  title={Qwen2 technical report},
  author={Yang, An and Yang, Baosong and Hui, Binyuan and Zheng, Bo and Yu, Bowen and Zhou, Chang and Li, Chengpeng and Li, Chengyuan and Liu, Dayiheng and Huang, Fei and others},
  journal={arXiv preprint arXiv:2407.10671},
  year={2024}
}

@article{gupta2024rebuilding,
  title={Rebuilding rome: Resolving model collapse during sequential model editing},
  author={Gupta, Akshat and Baskaran, Sidharth and Anumanchipalli, Gopala},
  journal={arXiv preprint arXiv:2403.07175},
  year={2024}
}

@misc{meng2023locatingeditingfactualassociations,
      title={Locating and Editing Factual Associations in GPT}, 
      author={Kevin Meng and David Bau and Alex Andonian and Yonatan Belinkov},
      year={2023},
      eprint={2202.05262},
      archivePrefix={arXiv},
      primaryClass={cs.CL},
      url={https://arxiv.org/abs/2202.05262}, 
}

@misc{chen2024extroversionintroversioncontrollingpersonality,
      title={Extroversion or Introversion? Controlling The Personality of Your Large Language Models}, 
      author={Yanquan Chen and Zhen Wu and Junjie Guo and Shujian Huang and Xinyu Dai},
      year={2024},
      eprint={2406.04583},
      archivePrefix={arXiv},
      primaryClass={cs.CL},
      url={https://arxiv.org/abs/2406.04583}, 
}

@misc{mao2024editingpersonalitylargelanguage,
      title={Editing Personality for Large Language Models}, 
      author={Shengyu Mao and Xiaohan Wang and Mengru Wang and Yong Jiang and Pengjun Xie and Fei Huang and Ningyu Zhang},
      year={2024},
      eprint={2310.02168},
      archivePrefix={arXiv},
      primaryClass={cs.CL},
      url={https://arxiv.org/abs/2310.02168}, 
}

@inproceedings{yang2021learning,
  title={Learning to answer psychological questionnaire for personality detection},
  author={Yang, Feifan and Yang, Tao and Quan, Xiaojun and Su, Qinliang},
  booktitle={Findings of the Association for Computational Linguistics: EMNLP 2021},
  pages={1131--1142},
  year={2021}
}

@article{chaplot2023albert,
  title={Albert q. jiang, alexandre sablayrolles, arthur mensch, chris bamford, devendra singh chaplot, diego de las casas, florian bressand, gianna lengyel, guillaume lample, lucile saulnier, l{\'e}lio renard lavaud, marie-anne lachaux, pierre stock, teven le scao, thibaut lavril, thomas wang, timoth{\'e}e lacroix, william el sayed},
  author={Chaplot, Devendra Singh},
  journal={arXiv preprint arXiv:2310.06825},
  year={2023}
}

@article{chen2021evaluating,
  title={Evaluating large language models trained on code},
  author={Chen, Mark and Tworek, Jerry and Jun, Heewoo and Yuan, Qiming and Pinto, Henrique Ponde De Oliveira and Kaplan, Jared and Edwards, Harri and Burda, Yuri and Joseph, Nicholas and Brockman, Greg and others},
  journal={arXiv preprint arXiv:2107.03374},
  year={2021}
}

@article{mccrae1992introduction,
  title={An introduction to the five-factor model and its applications},
  author={McCrae, Robert R and John, Oliver P},
  journal={Journal of personality},
  volume={60},
  number={2},
  pages={175--215},
  year={1992},
  publisher={Wiley Online Library}
}

@article{achiam2023gpt,
  title={Gpt-4 technical report},
  author={Achiam, Josh and Adler, Steven and Agarwal, Sandhini and Ahmad, Lama and Akkaya, Ilge and Aleman, Florencia Leoni and Almeida, Diogo and Altenschmidt, Janko and Altman, Sam and Anadkat, Shyamal and others},
  journal={arXiv preprint arXiv:2303.08774},
  year={2023}
}

@article{radford2019language,
  title={Language models are unsupervised multitask learners},
  author={Radford, Alec and Wu, Jeffrey and Child, Rewon and Luan, David and Amodei, Dario and Sutskever, Ilya and others},
  journal={OpenAI blog},
  volume={1},
  number={8},
  pages={9},
  year={2019}
}

@inproceedings{
mend_icrl2022,
title={Fast Model Editing at Scale},
author={Eric Mitchell and Charles Lin and Antoine Bosselut and Chelsea Finn and Christopher D Manning},
booktitle={International Conference on Learning Representations},
year={2022},
url={https://openreview.net/forum?id=0DcZxeWfOPt}
}

@inproceedings{serac_icrl2022,
    title={Memory-Based Model Editing at Scale},
    author={Mitchell, Eric and Lin, Charles and Bosselut, Antoine and Finn, Chelsea and Manning, Christopher D.},
    booktitle={International Conference on Machine Learning},
    url={https://arxiv.org/pdf/2206.06520.pdf},
    year={2022},
}

@inproceedings{llmKnolwedgeEditing,
  title={Editing Factual Knowledge in Language Models},
  author={De Cao, Nicola and Aziz, Wilker and Titov, Ivan},
  booktitle={Proceedings of the 2021 Conference on Empirical Methods in Natural Language Processing},
  pages={6491--6506},
  year={2021}
}

@article{ouyang2022training,
  title={Training language models to follow instructions with human feedback},
  author={Ouyang, Long and Wu, Jeffrey and Jiang, Xu and Almeida, Diogo and Wainwright, Carroll and Mishkin, Pamela and Zhang, Chong and Agarwal, Sandhini and Slama, Katarina and Ray, Alex and others},
  journal={Advances in neural information processing systems},
  volume={35},
  pages={27730--27744},
  year={2022}
}

@article{bai2022training,
  title={Training a helpful and harmless assistant with reinforcement learning from human feedback},
  author={Bai, Yuntao and Jones, Andy and Ndousse, Kamal and Askell, Amanda and Chen, Anna and DasSarma, Nova and Drain, Dawn and Fort, Stanislav and Ganguli, Deep and Henighan, Tom and others},
  journal={arXiv preprint arXiv:2204.05862},
  year={2022}
}

@article{ju2025probing,
  title={Probing then editing response personality of large language models},
  author={Ju, Tianjie and Shao, Zhenyu and Wang, Bowen and Chen, Yujia and Zhang, Zhuosheng and Fei, Hao and Lee, Mong-Li and Hsu, Wynne and Duan, Sufeng and Liu, Gongshen},
  journal={arXiv preprint arXiv:2504.10227},
  year={2025}
}

@article{karra2022estimating,
  title={Estimating the personality of white-box language models},
  author={Karra, Saketh Reddy and Nguyen, Son The and Tulabandhula, Theja},
  journal={arXiv preprint arXiv:2204.12000},
  year={2022}
}

@article{serapio2023personality,
  title={Personality traits in large language models},
  author={Serapio-Garc{\'\i}a, Gregory and Safdari, Mustafa and Crepy, Cl{\'e}ment and Sun, Luning and Fitz, Stephen and Abdulhai, Marwa and Faust, Aleksandra and Matari{\'c}, Maja},
  journal={arXiv preprint arXiv:2307.00184},
  year={2023}
}

@article{huang2023revisiting,
  title={Revisiting the reliability of psychological scales on large language models},
  author={Huang, Jen-tse and Jiao, Wenxiang and Lam, Man Ho and Li, Eric John and Wang, Wenxuan and Lyu, Michael R},
  journal={arXiv preprint arXiv:2305.19926},
  year={2023}
}

@article{wei2022emergent,
  title={Emergent abilities of large language models},
  author={Wei, Jason and Tay, Yi and Bommasani, Rishi and Raffel, Colin and Zoph, Barret and Borgeaud, Sebastian and Yogatama, Dani and Bosma, Maarten and Zhou, Denny and Metzler, Donald and others},
  journal={arXiv preprint arXiv:2206.07682},
  year={2022}
}

@article{liu2016language,
  title={A language-independent and compositional model for personality trait recognition from short texts},
  author={Liu, Fei and Perez, Julien and Nowson, Scott},
  journal={arXiv preprint arXiv:1610.04345},
  year={2016}
}

@inproceedings{vstajner2020survey,
  title={A survey of automatic personality detection from texts},
  author={{\v{S}}tajner, Sanja and Yenikent, Seren},
  booktitle={Proceedings of the 28th international conference on computational linguistics},
  pages={6284--6295},
  year={2020}
}

@article{vu2017lexical,
  title={Lexical-semantic resources: yet powerful resources for automatic personality classification},
  author={Vu, Xuan-Son and Flekova, Lucie and Jiang, Lili and Gurevych, Iryna},
  journal={arXiv preprint arXiv:1711.09824},
  year={2017}
}

@inproceedings{mairesse2006automatic,
  title={Automatic recognition of personality in conversation},
  author={Mairesse, Fran{\c{c}}ois and Walker, Marilyn},
  booktitle={Proceedings of the human language technology conference of the NAACL, companion volume: short papers},
  pages={85--88},
  year={2006}
}

@article{kampman2018investigating,
  title={Investigating audio, visual, and text fusion methods for end-to-end automatic personality prediction},
  author={Kampman, Onno and Barezi, Elham J and Bertero, Dario and Fung, Pascale},
  journal={arXiv preprint arXiv:1805.00705},
  year={2018}
}

@article{mccrae1989reinterpreting,
  title={Reinterpreting the Myers-Briggs type indicator from the perspective of the five-factor model of personality},
  author={McCrae, Robert R and Costa Jr, Paul T},
  journal={Journal of personality},
  volume={57},
  number={1},
  pages={17--40},
  year={1989},
  publisher={Wiley Online Library}
}

@article{pittenger1993utility,
  title={The utility of the Myers-Briggs type indicator},
  author={Pittenger, David J},
  journal={Review of educational research},
  volume={63},
  number={4},
  pages={467--488},
  year={1993},
  publisher={Sage Publications Sage CA: Thousand Oaks, CA}
}

@article{white2023prompt,
  title={A prompt pattern catalog to enhance prompt engineering with chatgpt},
  author={White, Jules and Fu, Quchen and Hays, Sam and Sandborn, Michael and Olea, Carlos and Gilbert, Henry and Elnashar, Ashraf and Spencer-Smith, Jesse and Schmidt, Douglas C},
  journal={arXiv preprint arXiv:2302.11382},
  year={2023}
}

@article{cui2023machine,
  title={Machine mindset: An mbti exploration of large language models},
  author={Cui, Jiaxi and Lv, Liuzhenghao and Wen, Jing and Wang, Rongsheng and Tang, Jing and Tian, Yonghong and Yuan, Li},
  journal={arXiv preprint arXiv:2312.12999},
  year={2023}
}

@article{du2005relation,
  title={The relation between personality type and sense of coherence among technical workers},
  author={Du Toit, F and Coetzee, S and Visser, D},
  journal={Southern African Business Review},
  volume={9},
  number={1},
  pages={51--65},
  year={2005},
  publisher={UNISA Press}
}

@article{mairesse2011controlling,
  title={Controlling user perceptions of linguistic style: Trainable generation of personality traits},
  author={Mairesse, Fran{\c{c}}ois and Walker, Marilyn A},
  journal={Computational Linguistics},
  volume={37},
  number={3},
  pages={455--488},
  year={2011},
  publisher={MIT Press One Rogers Street, Cambridge, MA 02142-1209, USA journals-info~…}
}

@article{ribeiro2020beyond,
  title={Beyond accuracy: Behavioral testing of NLP models with CheckList},
  author={Ribeiro, Marco Tulio and Wu, Tongshuang and Guestrin, Carlos and Singh, Sameer},
  journal={arXiv preprint arXiv:2005.04118},
  year={2020}
}

@article{benjamin2018redefine,
  title={Redefine statistical significance},
  author={Benjamin, Daniel J and Berger, James O and Johannesson, Magnus and Nosek, Brian A and Wagenmakers, E-J and Berk, Richard and Bollen, Kenneth A and Brembs, Bj{\"o}rn and Brown, Lawrence and Camerer, Colin and others},
  journal={Nature human behaviour},
  volume={2},
  number={1},
  pages={6--10},
  year={2018},
  publisher={Nature Publishing Group UK London}
}

@inproceedings{dror2018hitchhiker,
  title={The hitchhiker’s guide to testing statistical significance in natural language processing},
  author={Dror, Rotem and Baumer, Gili and Shlomov, Segev and Reichart, Roi},
  booktitle={Proceedings of the 56th Annual Meeting of the Association for Computational Linguistics (Volume 1: Long Papers)},
  pages={1383--1392},
  year={2018}
}

@article{artstein2008inter,
  title={Inter-coder agreement for computational linguistics},
  author={Artstein, Ron and Poesio, Massimo},
  journal={Computational linguistics},
  volume={34},
  number={4},
  pages={555--596},
  year={2008},
  publisher={MIT Press One Rogers Street, Cambridge, MA 02142-1209, USA journals-info~…}
}

@article{meng2022locating,
  title={Locating and editing factual associations in gpt},
  author={Meng, Kevin and Bau, David and Andonian, Alex and Belinkov, Yonatan},
  journal={Advances in neural information processing systems},
  volume={35},
  pages={17359--17372},
  year={2022}
}
\clearpage
\appendix


\section{Additional Experiments}
\subsection{Leveraging other Model Editing Methods}
\label{appendix:MEMIT}
In addition to \textbf{r-ROME}, we also experimented with another representative editing method, \textbf{MEMIT}. \textbf{MEMIT} supports \textit{multi-token, multi-fact editing} by distributing updates across multiple layers, which makes it effective for modifying distributed factual knowledge where consistent generalization across contexts is required.

As shown in Table~\ref{tab:mbti_model_results_MEMIT}, PALETTE\textsuperscript{MEMIT} generally shows smaller improvements in personality alignment 
compared to PALETTE\textsuperscript{r-ROME}. This difference can be explained from the perspective of generalization: 
MEMIT approach is effective for modifying distributed factual knowledge, where generalization involves consistently editing the same fact across various contexts. However, personality is inherently context-dependent, often expressed through diverse but semantically aligned utterances (e.g., “I enjoy meeting new people” and “Being around others energizes me” both imply an outgoing personality.). Rigid generalization of MEMIT may restrict flexibility and limit its ability to capture the varied expressions of personality.

For this reason, in our main experiments we primarily adopt r-ROME, which better accommodates the contextual variability inherent to personality editing.

\subsection{Analysis on Low Response Quality for Perceiving Personality}
\label{appendix:P}

To stated and demonstrated by the low effectiveness on the Perceiving (P) personality, we interpret this phenomenon as being due to a conflict between the linguistic expressions associated with the Perceiving tendency and the training distribution of typical models. The Perceiving dimension tends to favor more open and fluid expressions, whereas most LLMs are trained to generate clear and structured responses. This mismatch seems to make it difficult for the model to reproduce these uncertain linguistic styles after editing. 

\begin{table}[ht!]
\centering
\resizebox{\linewidth}{!}{
\begin{tabular}{lccccccccc}
\toprule
 \textbf{Evaluation} &\textbf{Base} & \textbf{E} & \textbf{I} & \textbf{N} & \textbf{S} & \textbf{F} & \textbf{T} & \textbf{P} & \textbf{J} \\
\midrule
                        
                           Naturalness  &4.08    &4.14       &3.99       &4.03       &4.07       &4.11       &4.14       &\textcolor{red}{3.64}       &4.12       \\
                           Coherence &4.06     &4.29       &4.16       &4.09       &4.13       &4.18      &4.42       &\textcolor{red}{3.78}       &4.05       \\
                           \cmidrule(lr){1-10} 
                           HumanEval  & 0.14     &0.12    &0.13   & 0.12       & 0.14       & 0.12     &0.12      &\textcolor{red}{0.10}       &0.14           \\
\bottomrule
\end{tabular}
}
\caption{PALETTE\textsuperscript{r-ROME} response quality results for Mistral-7B-instruct-v0.3 in MBTI (E/I, N/S, F/T, P/J).}
\label{tab:P}
\end{table}

To ensure this issue is not specific to the Qwen-2.5-1.5B model, we also conducted experiments using the Mistral-7B-Instruct-v0.3 model, as shown in Table~\ref{tab:P}. We observed similar performance degradation in both naturalness and coherence, indicating that the low reproducibility of the Perceiving (P) tendency is not an issue confined to a particular model but rather a common phenomenon closely related to the structural learning characteristics of LLMs.

\subsection{Adjusting Multiple MBTI Dimensions Simultaneously}
\label{appendix:multi-editing}
While our main experiments focus on adjusting one MBTI personality trait at a time (e.g., E/I, S/N), our assessment-based model editing approach is inherently scalable to multiple traits. To investigate this, we additionally conducted an experiment targeting both Introversion (I) and Thinking (T) traits simultaneously.

\begin{table}[ht!]
    \centering
    \small
    \resizebox{\linewidth}{!}{
    \begin{tabular}{
        p{4.0cm}
        >{\centering\arraybackslash}p{1.2cm}
        >{\centering\arraybackslash}p{1.2cm}
    }        \toprule
        \textbf{Setting}  & \textbf{I} & \textbf{T} \\
        \midrule
            Base                        &0.560 &0.365  \\

            \cmidrule(lr){1-3}
            
            I-PALETTE\textsuperscript{r-ROME} &-  &0.636   \\

            T-PALETTE\textsuperscript{r-ROME}                       &0.620 &-  \\
            \cmidrule(lr){1-3}
            I+T-PALETTE\textsuperscript{r-ROME}   &\textbf{0.665} &\textbf{0.652}   \\
        \bottomrule
    \end{tabular}
    }
    \caption{PALETTE\textsuperscript{r-ROME} target personality expression rate results in Qwen-2.5-1.5B for MBTI dimensions I and T personality. The best result is bolded.}
    \label{tab:multi-editing}
\end{table}

The results in Table~\ref{tab:multi-editing} demonstrate that PALETTE is scalable to adjusting multiple MBTI traits simultaneously, as shown by the significant adjustments in both the T and I traits.

\begin{table*}[ht!]
    \centering
    \small
    \resizebox{0.85\textwidth}{!}{
    \begin{tabular}{
        p{3.0cm}
        p{3.4cm}|
        >{\centering\arraybackslash}p{0.9cm}
        >{\centering\arraybackslash}p{0.9cm}|
        >{\centering\arraybackslash}p{0.9cm}
        >{\centering\arraybackslash}p{0.9cm}|
        >{\centering\arraybackslash}p{0.9cm}
        >{\centering\arraybackslash}p{0.9cm}|
        >{\centering\arraybackslash}p{0.9cm}
        >{\centering\arraybackslash}p{0.9cm}
    }        \toprule[1.3pt]
        \textbf{Model} & \textbf{Setting} & \textbf{E} & \textbf{I} & \textbf{N} & \textbf{S} & \textbf{F} & \textbf{T} & \textbf{P} & \textbf{J} \\
        \midrule[1.1pt]
        \multirow{6}{*}{Qwen-2.5-1.5B} 
            & Base                        &0.410 &0.560 &0.420  &0.580  &0.635  &0.365  &0.492  &0.508  \\
            & PALETTE\textsuperscript{MEMIT} &0.476  &0.573  &0.443  &0.521  &0.638  &0.450  &0.522  &0.486  \\
            & PALETTE\textsuperscript{r-ROME}                        &0.524 &\underline{0.636} &0.521  &\underline{{0.685}}  &0.726 
            &\underline{0.620}  &0.547  &\underline{0.634} \\
            \cmidrule(lr){2-10} 
            & Prompt   &\underline{0.716} &0.560 &\underline{0.756}  &0.630  &0.723  &0.305  &0.578  &0.549  \\
            
            & PALETTE\textsuperscript{MEMIT} w/ prompt &0.715  &0.589  &0.732  &0.623  &\underline{0.735}  &0.440  &\underline{0.609}  &0.576  \\
            & PALETTE\textsuperscript{r-ROME}  w/ prompt   &\textbf{0.728}  &\textbf{0.685}  &\textbf{0.805}  &\textbf{0.728}  &\textbf{0.778}  &\textbf{0.665}  &\textbf{0.623}  &\textbf{0.648}  \\
        \specialrule{1.1pt}{3pt}{3pt}
        \multirow{6}{*}{Mistral-7B-Instruct-v0.3} 
            & Base                        &0.476  &0.524  &0.245  &0.755  &0.619  &0.381  &0.494  &0.506  \\
            & PALETTE\textsuperscript{MEMIT}   &0.475  &0.530  &0.355  &0.761 &0.627 &0.399  &0.497  &0.512  \\
            & PALETTE\textsuperscript{r-ROME}  &0.485  &0.585  &0.403  &0.780  &0.664  &0.444  &0.529  &0.545  \\
            \cmidrule(lr){2-10} 
            & Prompt                      &\underline{0.699}  &0.589  &\underline{0.823}  &0.794
            &\underline{0.786}  &0.585  &0.711  &0.780  \\
            & PALETTE\textsuperscript{MEMIT} w/ prompt &0.678  &\underline{0.602}  &0.820  &\underline{0.802}  &0.778  &\underline{0.587}  &\underline{0.818}  &0.776  \\
            & PALETTE\textsuperscript{r-ROME} w/ prompt  &\textbf{0.711}  &\textbf{0.678}  
            &\textbf{0.826}  
            &\textbf{0.805}  &\textbf{0.791}  &\textbf{0.591}  &\textbf{0.845}  &\textbf{0.782}  \\
        \bottomrule[1.3pt]
    \end{tabular}
    }
    \caption{Target personality expression rate results in Qwen-2.5-1.5B and Mistral-7B-Instruct-v0.3 for MBTI dimensions (I/E, N/S, F/T, P/J) leveraging and comparing with MEMIT. The best result is bolded, and the second-best is underlined.}
    \label{tab:mbti_model_results_MEMIT}
\end{table*}

\subsection{Few-shot Prompt Baseline Using Self-Referential Demonstrations}
\label{appendix:fewshot-prompting}

We additionally evaluate a Few-shot Prompt setting that uses the same 12 self-referential personality queries employed in PALETTE as in-context demonstrations.

While our original prompt baseline already includes explicit natural-language descriptions for each personality dimension, this few-shot configuration provides a more direct comparison by embedding the identical 12 items directly into the input prompt. In this setup, each query–response pair is formatted as an in-context example, and the target query is answered by the model conditioned on these demonstrations.

\begin{table*}[ht!]
    \centering
    \small
    \resizebox{\linewidth}{!}{
    \begin{tabular}{
        p{3.0cm}
        p{3.7cm}|
        >{\centering\arraybackslash}p{0.9cm}
        >{\centering\arraybackslash}p{0.9cm}|
        >{\centering\arraybackslash}p{0.9cm}
        >{\centering\arraybackslash}p{0.9cm}|
        >{\centering\arraybackslash}p{0.9cm}
        >{\centering\arraybackslash}p{0.9cm}|
        >{\centering\arraybackslash}p{0.9cm}
        >{\centering\arraybackslash}p{0.9cm}
    }
        \toprule[1.3pt]
        \textbf{Model} 
        & \textbf{Setting} 
        & \textbf{E} & \textbf{I} 
        & \textbf{N} & \textbf{S} 
        & \textbf{F} & \textbf{T} 
        & \textbf{P} & \textbf{J} \\
        \midrule[1.1pt]

        \multirow{4}{*}{Qwen-2.5-1.5B}
            & Base 
            & 0.410 & 0.560 & 0.420 & 0.580 & 0.635 & 0.365 & 0.492 & 0.508 \\

            & \textit{Few-shot Prompt}
            & 0.475 & 0.589 & 0.510 & 0.512 & 0.639 & 0.430 & 0.520 & 0.503 \\

            & \textbf{PALETTE}
            & \underline{0.524} & \underline{0.636} & \underline{0.521} & \underline{0.685} & \underline{0.726} & \underline{0.620} & \underline{0.547} & \underline{0.634} \\

            & \textbf{PALETTE} w/ prompt
            & \textbf{0.728} & \textbf{0.685} & \textbf{0.805} & \textbf{0.728}
            & \textbf{0.778} & \textbf{0.665} & \textbf{0.623} & \textbf{0.648} \\

        \bottomrule[1.3pt]
    \end{tabular}
    }
    \caption{
    Few-shot prompting baseline using the same 12 self-referential personality queries as in-context demonstrations.
    The best result is bolded and the second-best is underlined.
    }
    \label{tab:fewshot-prompting}
\end{table*}

Table~\ref{tab:fewshot-prompting} reports the results on Qwen-2.5-1.5B under four settings: the base model, PALETTE, Few-shot Prompting, and PALETTE combined with prompting.

As shown in the results, Few-shot Prompting yields a modest improvement over the base model across several traits, indicating that the 12 self-referential queries themselves contain meaningful personality signals that can be partially exploited through in-context learning. However, its performance remains substantially below that of PALETTE across all eight personality dimensions.

Notably, PALETTE consistently outperforms the Few-shot Prompt baseline despite using the same underlying queries, demonstrating that the observed gains cannot be attributed solely to exposure to demonstrations. Furthermore, combining PALETTE with prompting achieves the strongest performance, suggesting that prompt-based methods and model editing are complementary: prompting offers flexible short-term conditioning, while PALETTE provides persistent and structurally grounded personality control.

These results confirm that PALETTE introduces benefits beyond what few-shot prompting alone can achieve, validating the necessity of model-level personality editing.

\section{Prompts}
\label{appendix:prompts}

\subsection{Response Generation Prompts}
\label{appendix:tailored_prompt}
We design and use BASE prompt, Personality-inducing prompt as shown in Table~\ref{tab:generation_prompt}.
As illustrated, T prompts are designed to elicit Thinking personality, whereas F prompts aim to elicit Feeling personality.  

\subsection{Personality Editing Evaluation Prompts}
\label{appendix:editing_prompt}
For target/opposing personality comparison evaluation, we conduct pairwise comparisons between PALETTE and Base model based on alignment with the target personality. Example Prompt can be seen in Table~\ref{tab:evaluation_prompt}. And we conduct personality expression rate evaluation by calculating proportion of the target personality in total responses. This prompt example is mentioned in Table~\ref{tab:evaluation_prompt_ratio}.

\subsection{Response Quality Evaluation Prompts}
In evaluating naturalness and coherence, we employ prompting in gpt-4o, as illustrated in Table~\ref{tab:model_eval_prompt}.

\section{Adjustment Queries}
\label{appendix:request_prompts}

\subsection{Construction of Assessment Queries}
\label{appendix:assessment_queries}

Our assessment queries are constructed based on the MBTI questionnaire from 16Personalities\footnote{\url{{https://www.16personalities.com/free-personality-test}}}. 
Each query targets one of the four MBTI dimensions. We carefully rephrased each question into declarative statements suitable for representing both opposing traits, while ensuring semantic equivalence with the original items. 

\subsection{Example Adjustment Queries}
\label{appendix:total_adjustment}

We provide 3 examples for each MBTI dimension adjustment queries in Table~\ref{tab:total_request_E} to Table~\ref{tab:total_request_J}.

\section{Extra Implementation Details}
\paragraph{Computational Cost}
\label{appendix:cost}
Applying \textsc{PALETTE} to the \textit{Qwen-1.5B} model once with 12 queries takes approximately 26.80 seconds on an RTX A6000. Although this small computational overhead occurs, \textsc{PALETTE} enables inference efficiency and consistent personality expression by directly embedding modifications. In contrast, full \textsc{QLoRA-SFT} training requires substantially higher computational cost, with an average training runtime of 634.92 seconds even under identical hardware conditions. This comparison highlights that \textsc{PALETTE} achieves comparable or superior controllability at less than 5\% of the training time and with only 12 editing samples, demonstrating its remarkable data and compute efficiency.

\paragraph{SFT Extra Implementation}
\label{appendix:sft}

For fine-tuning baselines, we employ the \texttt{Machine\_Mindset\_MBTI} dataset\footnote{\url{https://huggingface.co/datasets/pandalla/Machine_Mindset_MBTI_dataset}}, which provides labeled utterances across MBTI personality dimensions. We split the dataset into training and validation sets for each dimension (e.g., F/T, E/I, J/P, N/S), with approximately $10$k--$23$k training samples and $2$k--$23$k validation samples depending on the trait. Detailed Configuration can be found in Table~\ref{tab:sft_config}.

\paragraph{PALETTE Hyper-parameter Adjustment}
\label{appendix:extra_implementation}
To adapt the r-ROME framework for personality editing,  several key hyperparameters were adjusted from the original GPT-2-XL configuration as shown in Table~\ref{tab:model_config}.

These changes optimize the model's ability to express nuanced personality types while aligning with the \textit{Qwen} model’s architecture.

\section{Human Evaluation Details}
\label{appendix:human_eval}

To assess the effectiveness of our personality editing approach, we conduct human evaluations using a structured assessment sheet, as shown in Table~\ref{tab:human-eval-guide}. We recruited three fluent English-speaking judges for the evaluation, each compensated at approximately \$10 per hour. Three judges were provided with an explanation of the personality traits, along with the speaker's utterance and model's responses, allowing them to compare personality before and after editing. We measured effectiveness using the win/lose ratio. Fleiss' kappa scores were 0.67. These results support the reliability of our human evaluations while maintaining independent judgment.

\begin{table*}[t]
\centering
\scalebox{0.85}{
\begin{tabularx}{\textwidth}{X} 
\hline
\hline
\\
\textbf{[BASE Prompt]} \\
{[Instruction]}\\
You are engaging a conversation with a human.
ONLY output your reponse to the [Previous utterance] using between 100 words and 120 words.
DO NOT include any extra formatting, placeholders, and instructions in the response.\\ 
\\
{[Previous utterance]} \\
\textit{speaker's previous utterance inserted} \\
\\
{[Your response]} \\ 
\\
\hline
\\
\textbf{[T Prompt]} \\
{[Instruction]}\\
You are engaging a conversation with a human.
Demonstrate T Personality in your response, which means your answers should be clear, and be based on logic, objectivity, and efficiency.
ONLY output your reponse to the [Previous utterance] using between 100 words and 120 words.
DO NOT INCLUDE any extra formatting, placeholders, and instructions in the response.
DO NOT MENTION T Personality in your response in any way.\\
\\      
{[Previous utterance]} \\
\textit{speaker's previous utterance inserted}\\
\\
{[Your response]} \\ 
\\
\hline
\\
\textbf{[F Prompt]} \\
{[Instruction]}\\
You are engaging a conversation with a human.
Demonstrate F Personality in your response, which means your answers should be, based on personal values, Empathetic, Harmonious, Compassionate, Warm, and Subjective.
ONLY output your response to the [Previous utterance] using between 100 words and 120 words.
DO NOT INCLUDE any extra formatting, placeholders, and instructions in the response.
DO NOT MENTION F Personality in your response in any way.\\
\\
{[Previous utterance]} \\
\textit{speaker's previous utterance inserted} \\
\\
{[Your response]} \\
\\
\hline
\hline
\end{tabularx}}
\caption{Example for response generation prompt (Base and \textit{Nature} dimension).}
\label{tab:generation_prompt}
\end{table*}
\begin{table*}[t]
\centering
\scalebox{0.85}{
\begin{tabularx}{\textwidth}{X}
\hline
\hline
\\
\textbf{[Target Personality Alignment Comparison Evaluation Prompt]} \\\\
You are an expert in Psychometrics, especially 16 Personality, Decision-Making Preferences dimension.\\
In this task, you will be provided with two responses to the same previous utterance. Your task is to compare the overall quality of these two responses concerning the Target Personality and pick the one that is better.\\
\\                        
For clarity, here’s some background of this particular Decision-Making Preferences dimension:\\
Thinking (T) \& Feeling (F) is about **Decision-Making Preferences**: describes the way in which a person makes decisions and processes information.\\
\\
Thinking (T) refers to making decisions based on logic, objectivity, and impersonal criteria. Thinkers prioritize truth, fairness, and consistency. They tend to be analytical, critical, and task-oriented. Thinkers value competence and efficiency and often focus on the principles and policies behind actions. They are Logical, Objective, Critical, Analytical, and Detached. \\
Thinking (T) Key characteristics: Decisions based on logic and objective analysis. \\
\\                        
Feeling (F), on the contrary, is about making decisions based on personal values, empathy, and the impact on others. Feelers prioritize harmony, compassion, and relationships. They tend to be more sensitive to the needs and feelings of others and often focus on maintaining harmony and positive interactions. Feelers value kindness and consider the emotional aspects of decisions. They are Empathetic, Harmonious, Compassionate, Warm, and Subjective. \\
Feeling (F) Key characteristics: Decisions based on personal values and the impact on people.\\
\\
{[Target Personality]} \\
\textit{target personality} \\
\\
{[Previous utterance]} \\
\textit{speaker's previous utterance inserted} \\
\\
{[Response 1]} \\
\textit{response 1} \\
\\                        
{[Response 2]}\\
\textit{response 2} \\
\\
{[Instruction]} \\
Compare the overall quality of these two responses and pick the one that is better at representing the Target Personality.\\
Please output in just following format: {\{"analysis": "Your analysis here.", "result": "1 or 2",\}} \\
e.g. {\{"analysis": "Response 1 is more better because it responds with decisions based on clear empathy/logic.", "result": "1",\}} \\
Don't explain why.\\
\\
\hline
\hline
\end{tabularx}}
\caption{Example for prompt in target personality alignment comparison evaluation.}
\label{tab:evaluation_prompt}
\end{table*}
\begin{table*}[t]
\centering
\scalebox{0.85}{
\begin{tabularx}{\textwidth}{X}
\hline
\hline
\\
\textbf{[Target Personality Expression Rate Evaluation Prompt]} \\\\
You are an expert in Psychometrics, especially 16 Personality. I am conducting the 16 Personality test on someone. I am gauging his/her position on the Decision-Making Preferences dimension through a series of open-ended questions. For clarity, here’s some background of this particular dimension:
\\
===\\
\\
Thinking (T) \& Feeling (F) is about **Decision-Making Preferences**: describes the way in which a person makes decisions and processes information. \\
\\
Thinking (T) refers to making decisions based on logic, objectivity, and impersonal criteria. Thinkers prioritize truth, fairness, and consistency. They tend to be analytical, critical, and task-oriented. Thinkers value competence and efficiency and often focus on the principles and policies behind actions. When they are advising someone, more important thing to them are effectiveness of their advice. They are Logical, Objective, Critical, Analytical, and Detached. \\
Key characteristics: Decisions based on logic and objective analysis.\\
\\
Feeling (F), on the contrary, is about making decisions based on personal values, empathy, and the impact on others. Feelers prioritize harmony, compassion, and relationships. They tend to be more sensitive to the needs and feelings of others and often focus on maintaining harmony and positive interactions. Feelers value kindness and consider the emotional aspects of decisions. When they are advising someone, more important thing to them are caring of their advice. They are Empathetic, Harmonious, Compassionate, Warm, and Subjective. \\
Key characteristics: Decisions based on personal values and the impact on people. \\
\\
=== \\
\\
My name is A. I’ve invited a participant B. I will input the conversations. \\
\\
Conversations: \\
A : \textit{speaker's previous utterance inserted}

B : \textit{LLM's response inserted}\\\\

Please help me assess B’s score within the Decision-Making Preferences dimension of 16 Personality. \\
You should provide the percentage of each category, which sums to 100\%, e.g., 30\% and 70\%. Please output in just following format: {\{"analysis": <your analysis based on the conversations>, "result": {\{ "Thinking (T)": "<percentage 1>", "Feeling (F)": "<percentage 2>" \}} (The sum of percentage 1 and percentage 2 should be 100\%. Output with percent sign.) \}}
e.g. {\{"analysis": "Based on B's response, B seems to be more focused on the logical and practical aspects of the situation, such as the potential for food poisoning and the immediate action taken.", "result": {\{ "Thinking (T)": "70\%", "Feeling (F)": "30\%" \}}\}}
Don't explain why. \\
\\
\hline
\hline
\end{tabularx}}
\caption{Example prompt of target personality expression rate evaluation in \textit{Thinking} Dimension.}
\label{tab:evaluation_prompt_ratio}
\end{table*}
\FloatBarrier
\begin{table*}[p]
\centering
\scalebox{0.85}{
\begin{tabularx}{\textwidth}{X}
\hline
\hline
\\
\textbf{[Response Quality Evaluation Prompt]} \\\\

You are tasked with evaluating the quality of dialogue focusing on conversational fluency and logical structure.\\
Use the following evaluation criteria regarding participant's Induced Personality:\\\\
1. Naturalness (1-5): How natural/fluent are the response?\\
2. Coherence (1-5): How logically consistent and coherent are the response?\\\\
===\\\\
My name is A. I’ve invited a participant B. I will input the conversations.\\\\
Conversations:\\\\
 A :  \textit{speaker's previous utterance inserted}\\
B :  \textit{response inserted}\\\\
In the provided conversations, evaluate B’s response by assigning scores for the two criteria (Naturalness, Coherence). Provide comments on samples that didn’t receive full marks.\\\\
B's Induced Personality : \textit{Edited Personality}\\\\
Please output in just following format: {{"Naturalness": <Naturalness score>, "Coherence": <Naturalness score>, "comment": <The reason why it did not receive full marks>}}\\\\
e.g.{{"Naturalness": 5, "Coherence": 4.5, "comment": "B's response was a little outside the subject."}}
Don't explain why.\\

\\
\hline
\hline
\end{tabularx}}
\caption{Example prompt for naturalness and coherence evaluation.}
\label{tab:model_eval_prompt}
\end{table*}
\FloatBarrier
\begin{table*}[t]
\centering
\scalebox{0.65}{
\begin{tabularx}{\textwidth}{X} 
\hline
\hline
\\
\{"prompt": "[Question] When meeting someone interesting, which do you usually find yourself: comfortably initiating a conversation, or hesitating to approach? Answer in one sentence. [Your answer] When meeting someone interesting, \{\} usually find myself", "subject": "I", "target\_new": \{"str": "comfortably"\}, \} \\
\\
\hline
\\
\{"prompt": "[Question] When choosing how to spend your time, which do you prefer: group-based activities, or solitary hobbies? Answer in one sentence. [Your answer] When choosing how to spend my time, \{\} prefer", "subject": "I", "target\_new": \{"str": "group"\} \} \\
\\
\hline
\\
\{"prompt": "[Question] At social gatherings, which do you usually find yourself: taking the lead, or waiting for others to initiate introductions? Answer in one sentence. [Your answer] At social gatherings, \{\} usually find myself", "subject": "I", "target\_new": \{"str": "taking"\} \} \\\\
\hline
\hline
\end{tabularx}}
\caption{Example adjustment queries used for editing E personality in \textit{Qwen-2.5-1.5b-inst.}}
\label{tab:total_request_E}
\end{table*}

\begin{table*}[ht]
\centering
\scalebox{0.65}{
\begin{tabularx}{\textwidth}{X} 
\hline
\hline
\\
\{"prompt": "[Question] Which do you usually find yourself: making new friends often, or sticking with your existing circle? Answer in one sentence. [Your answer] \{\} usually find myself", "subject": "I", "target\_new": \{"str": "sticking"\}, \} \\
\\
\hline
\\
\{"prompt": "[Question] When networking with strangers, which do you usually find yourself: daunted, or excited? Answer in one sentence. [Your answer] When networking with strangers, \{\} usually find myself", "subject": "I", "target\_new": \{"str": "daunted"\} \} \\
\\
\hline
\\
\{"prompt": "[Question] When choosing how to spend your time, which do you prefer: solitary hobbies, or group-based activities? Answer in one sentence. [Your answer] When choosing how to spend my time, \{\} prefer", "subject": "I", "target\_new": \{"str": "solitary"\} \} \\\\
\hline
\hline
\end{tabularx}}
\caption{Example adjustment queries used for editing I personality in \textit{Qwen-2.5-1.5b-inst.}}
\label{tab:total_request_I}
\end{table*}

\begin{table*}[ht]
\centering
\scalebox{0.65}{
\begin{tabularx}{\textwidth}{X} 
\hline
\hline
\\
\{"prompt": "[Question] When solving problems, which do you usually find yourself: experimenting with untested approaches, or relying on proven methods? Answer in one sentence. [Your answer] When solving problems, \{\} usually find myself", "subject": "I", "target\_new": \{"str": "experimenting"\}, \} \\
\\
\hline
\\
\{"prompt": "[Question] In conversations, which topic do you usually find more preferring: ethical dilemmas, or practical topics? Answer in one sentence. [Your answer] In conversations, topic \{\} usually find more preferring is", "subject": "I", "target\_new": \{"str": "ethical"\} \} \\
\\
\hline
\\
\{"prompt": "[Question] When thinking about the future, which do you usually find yourself focused on: imaginative theories and possibilities, or realistic trends and current facts? Answer in one sentence. [Your answer] When thinking about the future, \{\} usually find myself focused on", "subject": "I", "target\_new": \{"str": "imaginative"\} \} \\\\
\hline
\hline
\end{tabularx}}
\caption{Example adjustment queries used for editing N personality in \textit{Qwen-2.5-1.5b-inst.}}
\label{tab:total_request_N}
\end{table*}

\begin{table*}[ht]
\centering
\scalebox{0.65}{
\begin{tabularx}{\textwidth}{X} 
\hline
\hline
\\
\{"prompt": "[Question] When encountering new ideas, which do you usually find yourself more interested in: complex and novel concepts, or simple and straightforward ones? Answer in one sentence. [Your answer] \{\} usually find myself more interested in", "subject": "I", "target\_new": \{"str": "simple"\}, \} \\
\\
\hline
\\
\{"prompt": "[Question] In creative discussions, which do you usually find yourself focusing more: on concrete facts and outcomes, or on various interpretations with interest? Answer in one sentence. [Your answer] \{\} usually find myself focusing more on", "subject": "me", "target\_new": \{"str": "concrete"\} \} \\
\\
\hline
\\
\{"prompt": "[Question] When it comes to learning, which do you usually find yourself: seeking new experiences and areas of knowledge, or staying within familiar territory? Answer in one sentence. [Your answer] When it comes to learning, \{\} usually find myself", "subject": "me", "target\_new": \{"str": "staying"\} \} \\\\
\hline
\hline
\end{tabularx}}
\caption{Example adjustment queries used for editing S personality in \textit{Qwen-2.5-1.5b-inst.}}
\label{tab:total_request_S}
\end{table*}

\begin{table*}[t]
\centering
\scalebox{0.65}{
\begin{tabularx}{\textwidth}{X} 
\hline
\hline
\\
\{"prompt": "[Question] Which do you prioritize when determining a course of action: \{\} or people's feelings? Answer in one sentence. [Your answer] When determining a course of action, I prioritize", "subject": "facts", "target\_new": \{"str": "people"\} \} \\
\\
\hline
\\
\{"prompt": "[Question] Which do you favor in decisions, even if it means disregarding the other option: efficiency or emotional aspects? Answer in one sentence. [Your answer] In decisions, \{\} favor", "subject": "I", "target\_new": \{"str": "emotional"\} \} \\
\\
\hline
\\
\{"prompt": "[Question] When facts and feelings conflict, what do you usually find yourself following: your heart or your head? Answer in one sentence. [Your answer] When facts and feelings conflict, \{\} usually find myself following my", "subject": "I", "target\_new": \{"str": "heart"\} \} \\\\
\hline
\hline
\end{tabularx}}
\caption{Example adjustment queries used for editing F personality in \textit{Qwen-2.5-1.5b-inst.}}
\label{tab:total_request_F}
\end{table*}

\begin{table*}[ht]
\centering
\scalebox{0.65}{
\begin{tabularx}{\textwidth}{X} 
\hline
\hline
\\
\{"prompt": "[Question] Which do you \\ \qquad usually feel more persuaded by: emotionally resonating things with you, or by factual arguments? Answer in one sentence. [Your answer] \{\} usually feel more persuaded by", "subject": "I", "target\_new": \{"str": "factual"\}, \} \\
\\
\hline
\\
\{"prompt": "[Question] What speak louder to you: people's stories or numbers? Answer in one sentence. [Your answer] What speak louder to \{\} are", "subject": "me", "target\_new": \{"str": "numbers"\} \} \\
\\
\hline
\\
\{"prompt": "[Question] What speak louder to you: emotions or data? Answer in one sentence. [Your answer] What speak louder to \{\} are", "subject": "me", "target\_new": \{"str": "data"\} \} \\\\
\hline
\hline
\end{tabularx}}
\caption{Example adjustment queries used for editing T personality in \textit{Qwen-2.5-1.5b-inst.}}
\label{tab:total_request_T}
\end{table*}

\begin{table*}[ht]
\centering
\scalebox{0.65}{
\begin{tabularx}{\textwidth}{X} 
\hline
\hline
\\
\{"prompt": [Question] In your personal spaces, which do you usually find yourself: keeping things clean and organized, or letting things get a bit messy and unstructured? Answer in one sentence. [Your answer] In my personal spaces, \{\} usually find myself", "subject": "I", "target\_new": \{"str": "letting"\}, \} \\
\\
\hline
\\
\{"prompt": "[Question] In managing your time, which do you usually find yourself: using tools like schedules and lists, or handling things more spontaneously? Answer in one sentence. [Your answer] In managing my time, \{\} usually find myself", "subject": "I", "target\_new": \{"str": "handling"\} \} \\
\\
\hline
\\
\{"prompt": "[Question] At home, which do you usually find yourself: cleaning as soon as things get messy, or tolerating some mess for a while? Answer in one sentence. [Your answer] At home, \{\} usually find myself", "subject": "I", "target\_new": \{"str": "tolerating"\} \} \\\\
\hline
\hline
\end{tabularx}}
\caption{Example adjustment queries used for editing P personality in \textit{Qwen-2.5-1.5b-inst.}}
\label{tab:total_request_P}
\end{table*}

\begin{table*}[ht]
\centering
\scalebox{0.65}{
\begin{tabularx}{\textwidth}{X} 
\hline
\hline
\\
\{"prompt": "[Question] In your work or study life, which do you usually find yourself: maintaining a consistent schedule, or struggling to stick to schedule? Answer in one sentence. [Your answer] In your work or study life, \{\} usually find myself", "subject": "I", "target\_new": \{"str": "maintaining"\}, \} \\
\\
\hline
\\
\{"prompt": "[Question] When starting your day, which do you usually find yourself: making a to-do list, or going with the flow? Answer in one sentence. [Your answer] When starting your day, \{\} usually find myself", "subject": "I", "target\_new": \{"str": "making"\} \} \\
\\
\hline
\\
\{"prompt": "[Question] In uncertain situations, which do you usually find yourself: preferring clear direction, or adapting as things go? Answer in one sentence. [Your answer] In uncertain situations, \{\} usually find myself", "subject": "I", "target\_new": \{"str": "preferring"\} \} \\\\
\hline
\hline
\end{tabularx}}
\caption{Example adjustment queries used for editing J personality in \textit{Qwen-2.5-1.5b-inst.}}
\label{tab:total_request_J}
\end{table*}

\begin{table*}[ht]
\centering
\small
\sisetup{table-format=1.4, separate-uncertainty}
\renewcommand{\arraystretch}{1.2}
\begin{tabular}{l|l}
\toprule
\textbf{Parameter} & \textbf{Value} \\ 
\midrule
\textbf{Finetuning Type} & LoRA (rank = 32, target = all) \\ 
\textbf{Training Stage} & SFT (Supervised Fine-Tuning) \\ 
\textbf{Cutoff Length} & 4000 tokens \\ 
\textbf{Batch Size / Accum. Steps} & 4 / 8 \\ 
\textbf{Learning Rate} & 1e-4 \\ 
\textbf{Scheduler / Warmup Ratio} & Cosine / 0.1 \\ 
\textbf{Epochs} & 3.0 \\ 
\textbf{Precision} & bfloat16 \\ 
\textbf{Deepspeed Config} & ds\_z2\_config.json \\ 
\textbf{Evaluation Strategy} & steps (every 200 steps) \\ 
\textbf{Metric for Best Model} & \texttt{eval\_rouge-l} \\ 
\textbf{Generation Settings} & beams = 1, max\_length = 256 \\ 
\bottomrule
\end{tabular}
\caption{Configuration parameters for SFT in \textit{Qwen-2.5-1.5b-inst.}}
\label{tab:sft_config}
\end{table*}

\begin{table*}[ht]
\centering
\small
\sisetup{table-format=1.4, separate-uncertainty}
\renewcommand{\arraystretch}{1.2} 
\begin{tabular}{l|l}
\toprule
\textbf{Parameter} & \textbf{Value} \\ 
\midrule
\textbf{layers} & [15] \\ 
\textbf{fact\_token} & subject\_first \\ 
\textbf{v\_num\_grad\_steps} & 25 \\ 
\textbf{v\_lr} & 4e-1 \\ 
\textbf{v\_loss\_layer} & 27 \\ 
\textbf{v\_weight\_decay} & 1e-4 \\ 
\textbf{clamp\_norm\_factor} & 4 \\ 
\textbf{kl\_factor} & 0.0625 \\ 
\textbf{mom2\_adjustment} & false \\ 
\textbf{context\_template\_length\_params} & [[5, 10], [10, 10]] \\ 
\textbf{rewrite\_module\_tmp} & "model.layers.{}.mlp.down\_proj" \\ 
\textbf{layer\_module\_tmp} & "model.layers.{}" \\ 
\textbf{mlp\_module\_tmp} & "model.layers.{}.mlp" \\ 
\textbf{attn\_module\_tmp} & "model.layers.{}.attention.o\_proj" \\ 
\textbf{ln\_f\_module} & "model.final\_layernorm" \\ 
\textbf{lm\_head\_module} & "lm\_head" \\ 
\textbf{mom2\_dataset} & "wikipedia" \\ 
\textbf{mom2\_n\_samples} & 20 \\ 
\textbf{mom2\_dtype} & "float32" \\
\bottomrule
\end{tabular}
\caption{Configuration parameters for personality editing in \textit{Qwen-2.5-1.5b-inst.}}
\label{tab:model_config}
\end{table*}


\begin{table*}[t]
\centering
\scalebox{0.85}{
\begin{tabularx}{\textwidth}{X} 
\hline
\hline
\\
\textbf{[Instruction]} \\
\\For clarity, here’s some background of this particular Decision-Making Preferences dimension:\\
Thinking (T) \& Feeling (F) is about **Decision-Making Preferences**: describes the way in which a person makes decisions and processes information.\\\\
\textbf{Thinking (T)} refers to making decisions based on logic, objectivity, and impersonal criteria.\\
Thinkers prioritize truth, fairness, and consistency. They tend to be analytical, critical, and task-oriented.\\
Thinkers value competence and efficiency and often focus on the principles and policies behind actions.\\
They are Logical, Objective, Critical, Analytical, and Detached.\\
Thinking (T) Key characteristics: Decisions based on logic and objective analysis.\\\\
\textbf{Feeling (F)}, on the contrary, is about making decisions based on personal values, empathy, and the impact on others.\\
Feelers prioritize harmony, compassion, and relationships.\\
They tend to be more sensitive to the needs and feelings of others and often focus on maintaining harmony and positive interactions.\\
Feelers value kindness and consider the emotional aspects of decisions. They are Empathetic, Harmonious, Compassionate, Warm, and Subjective.\\
Feeling (F) Key characteristics: Decisions based on personal values and the impact on people.\\ \\
\\
\hline
\\
\textbf{[Target Personality: \textit{target personality}]} \\
\\Compare the overall quality of these two responses and pick the one that is better at representing the Target Personality.\\\\

{[Previous utterance]} \\
\textit{previous utterance} \\
\\
\\
\hline
\\
\textbf{[Response 1]} \\\\
\textit{response 1}\\
\\
\hline
\\
\textbf{[Response 2]} \\\\
\textit{response 2}\\
\\
\hline
\hline
\end{tabularx}}
\caption{An example of a structured assessment sheet used for human evaluation for \textit{Nature(T/F)} Dimension.}
\label{tab:human-eval-guide}
\end{table*}
\end{document}